\documentclass{article}
\usepackage[a4paper, margin=0.8in]{geometry}
\usepackage{natbib}
\usepackage{microtype}
\usepackage{algorithm}
\usepackage{algorithmic}
\usepackage{graphicx}
\usepackage{bm}
\usepackage{subfigure}
\usepackage{booktabs} 
\usepackage{kky}
\usepackage{comment}
\usepackage{graphicx}
\usepackage{todonotes}
\graphicspath{{./figures/}}
\newcommand{\lambdab}{\bm{\lambda}}
\newcommand{\sigmab}{\bm{\sigma}}
\newcommand{\stil}{{\tilde{s}}}

\usepackage{hyperref}
\usepackage{lipsum}

\title{Adaptive Multi-View ICA: Estimation of noise levels for optimal inference}

\author{%
  Hugo Richard \\
  Inria, Université Paris-Saclay\\
  Saclay, France \\
  \texttt{hugo.richard@inria.fr} \\
  \and

  Pierre Ablin \\
  Département de Mathématiques et Applications\\
  Ecole Normale Supérieure \\
  Paris, France \\
  \texttt{pierre.ablin@ens.fr} \\
 \and
 Aapo Hyv{\"a}rinen \\
 Inria, Université Paris-Saclay, Saclay, France\\
 Department of Computer Science HIIT, University of Helsinki, Finland \\
 \texttt{aapo.hyvarinen@helsinki.fi} \\
 \and
 Alexandre Gramfort \\
 Inria, Université Paris-Saclay\\
 Saclay, France \\
 \texttt{alexandre.gramfort@inria.fr} \\
 \and
 Bertrand Thirion \\
 Inria, Université Paris-Saclay\\
 Saclay, France \\
 \texttt{bertrand.thirion@inria.fr} \\
}
\begin{document}

\maketitle

\begin{abstract}

We consider a multi-view learning problem known as group independent component
analysis (group ICA), where the goal is to recover shared independent sources from many views.
The statistical modeling of this problem requires to take noise into account.
When the model includes additive noise on the observations, the likelihood is
intractable.
By contrast, we propose Adaptive multiView ICA (AVICA), a noisy ICA model where each view is a linear mixture of shared independent sources with additive noise on the sources.
In this setting, the likelihood has a tractable expression, which enables either direct optimization of the log-likelihood using a quasi-Newton method, or generalized EM.
Importantly, we consider that the noise levels are also parameters that are learned from the data. 
This enables sources estimation with a closed-form Minimum Mean Squared Error (MMSE) estimator which weights each view according to its relative noise level.
On synthetic data, AVICA yields better sources estimates than other group ICA methods thanks to its explicit MMSE estimator.
On real magnetoencephalograpy (MEG) data, we provide evidence that the
decomposition is less sensitive to sampling noise and that the noise
variance estimates are biologically plausible.
Lastly, on functional magnetic resonance imaging (fMRI) data, AVICA
exhibits best performance in transferring information across views.

\end{abstract}

\section{Introduction}
Unsupervised multiview learning aims at extracting some shared information from a collection of datasets.
 In practice, each view contributes a different amount
 of information. 
 A canonical example is in neuroimaging, where one is interested in
 recovering a common source from a group of subjects performing the
 same task. 
 In this setting, a view is given by the brain recording ---e.g. magneto-encephalography (MEG) or functional magnetic resonance imaging (fMRI)--- of one subject of a group.
 However, the magnitude of the response differs across subjects, i.e. data exhibit random effects~\cite{penny2007random}. 
 Moreover, the noise level associated with head movements, breath or heart beats is
 subject-specific and also depends on the brain region considered ~\cite{liu2016noise}.

 Independent component analysis~\cite{comon1994independent, hyvarinen2000independent} (ICA) is a
 widely used technique to recover independent components from a given signal.
 ICA has many applications such as finance~\cite{chen2007portfolio},
 astronomy~\cite{maino2002all}, telecommunications~\cite{ristaniemi1999performance} or bioinformatics~\cite{liebermeister2002linear}.
When ICA is used in a neuroimaging context, the unsupervised multiview learning problem of recovering common sources from multiple (possibly noisy) mixtures is
 called \emph{group ICA}.
 Most popular group ICA methods, routinely used to extract shared responses from
 neuroimaging data, do not optimize a proper likelihood
 ~\cite{calhoun2001method, varoquaux2009canica}. 
 Other works~\cite{richard2020modeling, guo2008unified} have introduced well principled models.
 However, \citet{richard2020modeling} assume identical noise variance across
 sources and views and \citet{guo2008unified} assume identical noise covariance
 across views.
 Lastly, in noisy mixtures, the optimal sources estimates are not given simply by
inverting the mixing matrices~\cite{hyvarinen1998independent} and they are
often impossible to evaluate in closed-form.

 In this paper, we introduce AVICA, a principled method that models multiview data as a linear mixture of sources with additive Gaussian noise on the sources.
 Importantly, AVICA allows for different noise variances depending on the sources and the views.
We show that AVICA is identifiable, and it has a closed-form likelihood as well as a closed-form solution for the E-step in the Expectation-Maximization (EM) algorithm. 
We introduce 
two methods for maximizing the likelihood: direct likelihood
maximization via an alternating quasi-Newton method and a quasi-Newton
EM algorithm.
We derive a closed-form minimum mean square error (MMSE) estimate for the
 sources which allows for finer source estimation.
 Lastly, we benchmark AVICA on an extensive set of experiments involving
 synthetic and real data based on two different neuroimaging modalities.
 On MEG data, we show that the recovered sources are more stable across trials
 and that the estimated noise variances correlate with the recorded noise level
 without stimulus. On fMRI data, AVICA exhibits the best performance in predicting the data of a left-out subject.

\section{Adaptive multiViewICA (AVICA)}

\textbf{Notation} The absolute value of the determinant of a matrix $W$ is
$|W|$. The $\ell_2$ norm of a vector $\sbb$ is $\|\sbb\|$.
We define the $\ell_{\infty}$ norm of a matrix $G$ by $\|G\|_{\infty} = \max_{i, j} |G_{ij}|$. The probability density function of the normal distribution with mean $\mu$ and
variance $\sigma^2$ is denoted $x \mapsto \mathcal{N}(x; \mu, \sigma^2)$.

\subsection{Model and likelihood}
We assume that we have $m$ different \emph{views} associated with a common set of events.
AVICA models the data of each view as a weighted combination of noisy
sources. The sources are common to each view and constitute the \emph{shared
  sources}. Denoting $\xb^i \in \mathbb{R}^{k}$ a random vector of data for view $i$ and $\sbb
\in \mathbb{R}^k$ a random vector of shared sources,
the model reads:
\begin{align*}
  &\xb^i = A^i(\sbb + \nb^i), \enspace i=1,\dots, m \\
  &p(\sbb) = \prod_{j=1}^k \delta(s_j) \\
  &\nb^i \sim \mathcal{N}(0, \Sigma^i) \\
  &\Sigma^i =
             \diag \left(\frac{(\sigma_1)^2}{m ({\lambda_1^i})^2} \dots \frac{(\sigma_k)^2}{m ({\lambda_k^i})^2} \right)
  \\
  &\forall j, \sum_{i=1}^m ({\lambda_j^i})^2 = 1 \numberthis \label{eq:avica}
\end{align*}
where $A^i \in \mathbb{R}^{k \times k}$ are unknown \emph{mixing matrices} for each of the $m$ views, and $\nb^i$
correspond to some additive noises on the sources.
The $k$ shared sources $\sbb$ are assumed independent with density $\delta$.
Finally, the noise terms $\nb^1 \dots \nb^m$ are
assumed to be independent across the $m$ views, and independent from the shared sources.

Considering a Gaussian assumption on the noise of the source $j$, we parametrize
its variance as  $\Sigma^i_j = \frac{({\sigma_j})^2}{m ({\lambda_j^i})^2}$,
 where $({\sigma_j})^2$ is the
\emph{global noise level} of source $j$, and $({\lambda_j^i})^2$ is the
\emph{relative noise precision} for view $i$. 
$\Sigma^i$ models the deviation between
the best possible source estimate one can recover from view $i$ and the shared
sources.
For a given source $j$, the relative noise precisions
are normalized so that they sum to one, $\sum_i({\lambda_j^i})^2 = 1$. Intuitively, $({\lambda_j^i})^2$ quantifies the influence of view $i$ on source $j$.
The parametrization using $({\lambda_j^i})^2$ and $(\sigma_j)^2$ instead of $\Sigma^i_j$ yields no loss of generality and its interest will become clear later.

The noise model of AVICA is equivalent to having noise on the sensors with covariance $A^i
\Sigma^i \left(A^i\right)^{\top}$, where $\Sigma^i$ is a diagonal matrix. As the
noise variance is inferred from the data, it is a
generalization of~\cite{richard2020modeling} in which the noise level is fixed and constant over views and sources.
Indeed for each source $j$ and view $i$, the relative noise precisions
$({\lambda_j^i})^2$, the global noise level $(\sigma_j)^2$ and the mixing
matrices $A^i$ are estimated.

The following proposition states that the AVICA model has the same
permutation and scaling indeterminacies as standard single-view ICA (see
appendix~\ref{app:identifiability} for the proof).
\begin{proposition}[Identifiability of Adaptive multiView ICA]
\label{prop:identifiability}
Let $\xb^i, \enspace i=1\dots m$ be generated
from equation~\eqref{eq:avica} with parameters $(A^i, \Sigma^i)_{i=1}^m$ and density $\delta$. Assume
that there exist parameters $({A'}^i, {\Sigma'}^i)_{i=1}^m$ and density $\delta'$ such
that $\xb^i = {A'}^i(\sbb' + { \nb' }^i)$ with $p(\sbb') = \prod_{j=1}^k
\delta'(s'_j)$ and ${\nb'}^i_j \sim \mathcal{N}(0, { \Sigma' }^i)$  where ${ \Sigma' }^i$ is a positive
diagonal matrix. Then, there exists a scale and permutation matrix $P\in
\bbR^{k\times k}$ such that for all $i$, ${ A' }^i = A^i P$ and $\Sigma^i = P
{\Sigma^i}' P^T$. 
\end{proposition}

We propose a likelihood-based approach for learning. As shown in the following
derivations, we get a likelihood in closed-form.
Let us denote by $W^i = (A^i)^{-1}$ the \emph{unmixing matrices} and view the likelihood as a function of $W^i, \lambda^i_j, \sigma_j$. 
By integrating over the sources (see also appendix~\ref{appendix:integration}), the negative
log-likelihood reads (up to a constant):
\begin{align*} 
  &\loss(W^i, \lambda^i_j, \sigma_j) = -\log \left(\int_\sbb p(\xb|\sbb)p(\sbb) d\sbb \right)  \\
  &= \sum_{i=1}^m \left[-\log(|W^i|) +\frac1{2}\sum_{j=1}^k \log \left(\frac{(\sigma_j)^2}{({\lambda^i_j})^2 m}\right)\right] -\log (\mathcal{J}) \numberthis \label{eq:likelihood}
\end{align*}
with
\begin{align*}
\mathcal{J} =  \int_{\sbb}\exp\left(-\sum_{i=1}^m
  \sum_{j=1}^k\frac{({\lambda^i_j})^2m}{2(\sigma_j)^2} (y^i_j -
  s_j)^2\right)p(\sbb) d\sbb
\end{align*}
where $y^i_j = (W^i \xb^i)_j$.
This integral is a product of one dimensional integrals since the coefficients of $s$ are independent. We obtain (see appendix~\ref{appendix:integration}): \\
$-\log(\mathcal{J}) = \sum_{i, j} \frac{m}{2(\sigma_j)^2} ({\lambda^i_j})^2
(y^i_j - \tilde{s}_j)^2 + \sum_j f(\tilde{s}_j, \sigma_j)$
where  \\
$f(\tilde{s}_j, \sigma_j) = -\log \left(\int_z \exp \left(-\frac{m}{2(\sigma_j)^2} z^2
  \right) \delta(\tilde{s}_j-z) dz\right)$ \\
involves the source density $\delta$ convolved by a Gaussian kernel and $\tilde{s}_j =
\sum_i y^i_j ({\lambda^i_j})^2$ is a weighted average of unmixed data.

To make the computation of $f$ available in closed-form while assuming a
heavy-tailed distribution,
we postulate a particular density $\delta$ consisting of the following Gaussian mixture:
\begin{align}
  \delta(s_j) = \frac12\left(\mathcal{N}( s_j; 0, \frac12) + \frac12\mathcal{N}( s_j; 0, \frac{3}{2})\right) \enspace.
  \label{density}
\end{align}
Denoting \\
$\phi_j(s_j, \sigma_j) = - \log( \mathcal{N}( s_j; 0, \frac12 +
\frac{\sigma_j^2}{m}) + \mathcal{N}( s_j; 0, \frac32 + \frac{\sigma_j^2}{m}))$ \\
$f$ reads:
$f(s_j, \sigma_j) = -\frac12\log(\frac{(\sigma_j)^2}{m}) + \phi(s_j, \sigma_j)$. \\
The negative log-likelihood becomes
\begin{align*}
          &\loss(W^i, \lambda^i_j, \sigma_j) = 
          \sum_{i=1}^m \left[- \log(|W^i|) - \sum_{j=1}^k\frac12 \log(({\lambda^i_j})^2) \right.\\
          &+ \left. \frac12 \sum_{j=1}^k \frac{m{\lambda^i_j }^2}{(\sigma_j)^2} \left(y^i_j - \sum_{z=1}^m y^z_j ({\lambda^z_j})^2 \right)^2 \right]\\
          &+ \sum_{j=1}^k \left[ \frac{1 - m}{2} \log\left(\frac{m}{(\sigma_j)^2}\right) + \phi \left(\sum_{z=1}^m y^z_j ({\lambda^z_j})^2, \sigma_j \right) \right] \\
          \numberthis \label{eq:avica_likelihood_unconstrained}
\end{align*}
where $\yb^i = W^i \xb^i$ and $\sum_{i=1}^m ({\lambda^i_j})^2 = 1$, for all $j$.
AVICA boils down to minimizing $\loss$.
When $\lambda^i_j = \frac{1}{\sqrt{m}}$ and $\sigma_j = 1$ we recover the
negative log-likelihood of MultiViewICA~\cite{richard2020modeling}. In
addition, if we only have one view
($m=1$), we recover the negative log-likelihood of
Infomax~\cite{bell1995information,cardoso1997infomax} where the density of the
source is
replaced by $\delta$ convolved with a Gaussian kernel.

                        \label{eq:avica_likelihood}

The following proposition shows that even if the data do not follow the density $\delta$, there exists a well-defined local minimum where the true
unmixing matrices are recovered up to some scaling.
We refer to appendix~\ref{app:robustness} for the proof.

\begin{proposition}[Robustness of AVICA w.r.t. source density misspecification]
  \label{prop:robustness}
  Consider $\xb^i, \enspace i=1\dots m,$ generated from equation~\eqref{eq:avica} with
  mixing matrices $A^i$, relative noise precisions ${\lambda^i_j}^*$ global noise ${\sigma_j^*}$ and
  a source density $\delta^*$ not necessarily equal to $\delta$ in equation~\ref{density}.

  We assume the data and the model verify the following hypothesis $(H)$:
  \begin{align}
    &\exists \mu > 0, \forall i, j \enspace ({\lambda^i_j})^2 \geq \mu^2, 
\end{align}

  There exist scaling matrices $\Gamma^i$, relative noise precisions $(\lambda^i_j)_{i=1,
    j=1}^{m, k}$ and global noise $(\sigma_j)_{j=1}^k$  such that $(\Gamma^i (A^i)^{-1})_{i=1}^m, (\lambda^i_j)_{i=1, j=1}^{m, k},
  (\sigma_j)_{j=1}^k$ is a local minimum of $\loss$~\eqref{eq:avica_likelihood_unconstrained}.
  This minimum is well-defined, meaning it does not occur at the border
  of the definition set of $\loss$~\eqref{eq:avica_likelihood_unconstrained}.
\end{proposition}
We include the constraints in $(H)$ in the AVICA model~\eqref{eq:avica} making $\mu^2$ a hyper-parameter fixed to $10^{-3}$ in all experiments.

\subsection{Efficient learning by quasi-Newton MLE} 
We first 
propose to optimize $\loss$ using an alternating quasi-Newton method we call
\emph{quasi-Newton MLE}. 
Quasi-Newton methods~\cite{nocedal2006numerical} minimize some function
iteratively by following the direction
$D = -\tilde{H}^{-1} G$ where $\tilde{H}^{-1}$ is an approximation of the inverse of the
Hessian  and $G$ is the gradient.

The proposed method is a block coordinate descent approach where we fix all
parameters except a specific group that undergoes a quasi-Newton step.
We cycle through the groups and repeat the procedure until the gradient norm is below a threshold, indicating that we are close to a stationary point.
\paragraph{Update w.r.t $W^i$}
The relative gradient $G^{W^i}\in\RR^{k\times k}$ and Hessian $H^{W^i} \in \mathbb{R}^{k, k, k, k}$ satisfy: 
$\loss((I + \epsilon) W^i) = \loss(W^i) + \langle \epsilon |G^{W^i} \rangle + \frac12 \langle
\epsilon |H^{W^i} \epsilon \rangle + o(\|\epsilon\|^2)$.
The Hessian is given by:
\begin{align*}
H^{W^i}_{abcd} &= \delta_{ad}\delta_{bc} +
                \delta_{ac}\left((\lambda^i_a)^4 \phi''(\tilde{s}_a) \right. \\
  & \left. + \frac{m}{\sigma_a^2}
  (1 - (\lambda^i_a)^2) (\lambda^i_a)^2\right)y^i_{b}y^i_d 
\end{align*}
for $a, b, c, d =1\dots k$, which is expensive to compute.
We use a similar approximation as in~\cite{ablin2018faster}, leading to the
approximation:
\begin{align*}
  \numberthis
  \label{Hessianw}
&\tilde{H}^{W^i}_{abcd} = \delta_{ad}\delta_{bc} +
\delta_{ac}\delta_{bd}\left((\lambda^i_a)^4 \phi''(\tilde{s}_a) + \right. \\
  & \left. \frac{m}{\sigma_a^2} (1 - (\lambda^i_a)^2)
  (\lambda^i_a)^2\right)(y^i_{b})^2
\end{align*}

This approximation is exact when the unmixed data $\yb^i$ are independent.
Therefore as we get closer to convergence, the quality of the approximation
improves.
Besides, the Hessian approximation is block-diagonal with blocks of size $2
\times 2$, making it easy to invert and regularize. Regularization is indeed
necessary to guarantee positivity of the Hessian and therefore a descent direction.
Updates are given by
\begin{equation}
  \label{update_W}
W^i
\leftarrow (I - \rho (\tilde{H}^{W^i})^{-1} G^{W^i}) W^i
\end{equation}
where $\rho \in \mathbb{R}$ is the stepsize.

The stepsize is chosen by backtracking line-search:
 we start with $\rho =1$ and halve the stepsize until it yields a decrease of the loss.

\paragraph{Update w.r.t $(\lambda^i_j)_{i=1}^m$}
The relative noise precisions parameters are subject to the two constraints:
$\sum_{i=1}^m (\lambda^i_j)^2 = 1$ and $\forall i \enspace (\lambda^i_j)^2 \geq \mu^2$.
To simplify optimization we make the following change of variable:
$({\eta^i_j})^2 +
\mu^2 = ({\lambda^i_j})^2$ and therefore $\etab_j = (\eta^1_j \dots \eta^m_j)$ is on the sphere of radius $1 -
m \mu^2$. Facing optimization on a manifold, we employ a Riemannian strategy based on projection~\cite{absil2012projection}.

Updates are given by (see Appendix~\ref{app:gradients}):
\begin{equation}
\label{update_l}
\etab_j \leftarrow \frac{\etab_j - \rho 
  (H^{\etab_j})^{-1} G^{\etab_j}}{\|\etab_j - \rho (H^{\etab_j})^{-1} G^{\etab_j}\|\sqrt{1
    - m\mu^2}}
\end{equation}
where $\rho$ is the step-size
chosen similarly by backtracking line-search and $G^{\etab_j}$ and $H^{\etab_j}$
are respectively the Riemannian gradient and Hessian with respect to $\etab_j$.

\paragraph{Update w.r.t $\sigma_j$}
As there are no constraints on the scalar $\sigma_j$, the Newton updates 
simply read:
\begin{equation}
  \label{update_sigma}
\sigma_j \leftarrow \sigma_j - \rho
\frac{G^{\sigma_j}}{H^{\sigma_j}}
\end{equation}
where $\rho$ is the step-size
again chosen by backtracking line-search.

We provide the formulas for the gradients and Hessians in appendix~\ref{app:gradients}.

\paragraph{Stopping criterion} While performing the alternating quasi-Newton algorithm, we monitor the norm of the gradients $\|G\|_{\infty}$ and use them to stop the convergence of AVICA.

The pseudo code for the proposed optimization algorithm is provided in
Algorithm~\ref{algo:avica}.

\begin{algorithm}[tb]
\caption{Optimization of AVICA via quasi-Newton MLE}
\begin{algorithmic}
\STATE {\bfseries Input} Dataset $(\xb^i)_{i=1}^m$,
initial values for $W^i$, $\lambda^i_j$, $\sigma_j$.
  Tolerance parameter $\varepsilon=10^{-3}$. Constraint parameter $\mu^2=10^{-3}$.

\STATE Compute $\eta^i_j$ from $\lambda^i_j$, Set $\mathrm{tol}=\infty$.
 \WHILE{$\mathrm{tol}>\varepsilon$}
   \STATE $\mathrm{tol} = 0$ 
   \FOR{$i=1\dots m$}
     \STATE Compute $(\tilde{H}^{W^i})^{-1} G^{W^i}$ using the Hessian
     approximation in equation~\eqref{Hessianw} and update $W^i$ using~\eqref{update_W} 
     \STATE Set $\mathrm{tol} = \max(\mathrm{tol}, \|G^{W^i}\|_{\infty})$
   \ENDFOR

   \FOR{$j=1\dots k$}
   \STATE Compute $(H^{\etab_j})^{-1} G^{\etab_j}$ and update $\etab_j$
   using~\eqref{update_l}
   \STATE $\mathrm{tol} = \max(\mathrm{tol}, \|G^{\etab_j}\|_{\infty})$
   \STATE Compute $(H^{\sigma_j})^{-1} G^{\sigma_j}$ and update
   $\sigma_j$ using~\eqref{update_sigma}
   \STATE Set $\mathrm{tol} = \max(\mathrm{tol}, \|G^{\sigma_j}\|_{\infty})$
   \ENDFOR
 \ENDWHILE
 \STATE Compute $\lambda^i_j$ from $\eta^i_j$
 \STATE { \bfseries Return} unmixing matrices $(W^i)_{i=1}^m$, precisions
 $(\lambda^i_j)_{i=1, j=1}^{m, k}$, global noise levels
 $(\sigma_j)_{j=1}^m$, sources $(\bbE[s_j|\xb])_{j=1}^m$ using equation~\eqref{mmse}.
\end{algorithmic}
\label{algo:avica}
\end{algorithm}

\subsection{A closed-form minimum mean square error (MMSE) estimator of sources}
Let $\hat{\sbb}$ an estimate of the sources $\sbb$ from observations $\xb$. A minimum mean square error (MMSE) estimate of $s$ minimizes
$\hat{\sbb} \rightarrow \mathbb{E}\left[\|\hat{\sbb} - \sbb\|^2\right]$. It is unique and given by $\hat{\sbb}= \mathbb{E}[\sbb|\xb]$.
We now derive a closed-form solution for $\hat{\sbb}$.

Following the computations in appendix~\ref{appendix:mmse_derivation} we can write:
\begin{align*}
  &p(\xb, \sbb) = p(\xb | \sbb) p(\sbb) = \prod_{i=1}^m p(\xb^i | \sbb) p(\sbb) \\
  &\propto \prod_{i=1}^m \prod_{j=1}^k \exp \left(-\frac{({\lambda^i_j})^2m}{2(\sigma_j)^2}(y^i_j - s_j)^2 \right)\delta(s_j) \\
                        &\propto \prod_{j=1}^k \exp \left(-\frac{m}{2(\sigma_j)^2} (\tilde{s}_j - s_j)^2 \right) \delta(s_j)
\end{align*}
where we leave out terms that do not depend on $s_j$.

Using our Gaussian mixture assumption on the sources we obtain after following
the derivations provided in appendix~\ref{appendix:mmse_derivation} :
\begin{align*}
  &p(\xb, \sbb) \propto \prod_{j=1}^k \sum_{\alpha \in \{\frac12, \frac32\}} \theta_{\alpha, j} \mathcal{N}\left(s_j; \tilde{\mu}_j, \tilde{\nu}_j\right)
\end{align*}
where $\theta_{\alpha, j} = \mathcal{N}\left(\tilde{s}_j; 0, \sqrt{\alpha +
    \frac{(\sigma_j)^2}{m}}\right)$, $\tilde{\mu}_j = \frac{m\alpha
  \tilde{s}_j}{m\alpha + (\sigma_j)^2}$ and $\tilde{\nu}_j =  \frac{(\sigma_j)^2 \alpha}{m \alpha + (\sigma_j)^2}$.
Therefore 
\begin{align*}
  &p(\sbb|\xb) = \frac{p(\xb, \sbb)}{p(\xb)} 
  = \prod_{j=1}^k \frac{\sum_{\alpha \in \{\frac12, \frac32\}}\theta_{\alpha, j} \mathcal{N}\left(s_j; \tilde{\mu}_j, \tilde{\nu}_j\right) }{\sum_{\alpha \in \{\frac12, \frac32\}}\theta_{\alpha, j}}
\end{align*}

\begin{figure}
    \centering
    \includegraphics[width=0.55\textwidth]{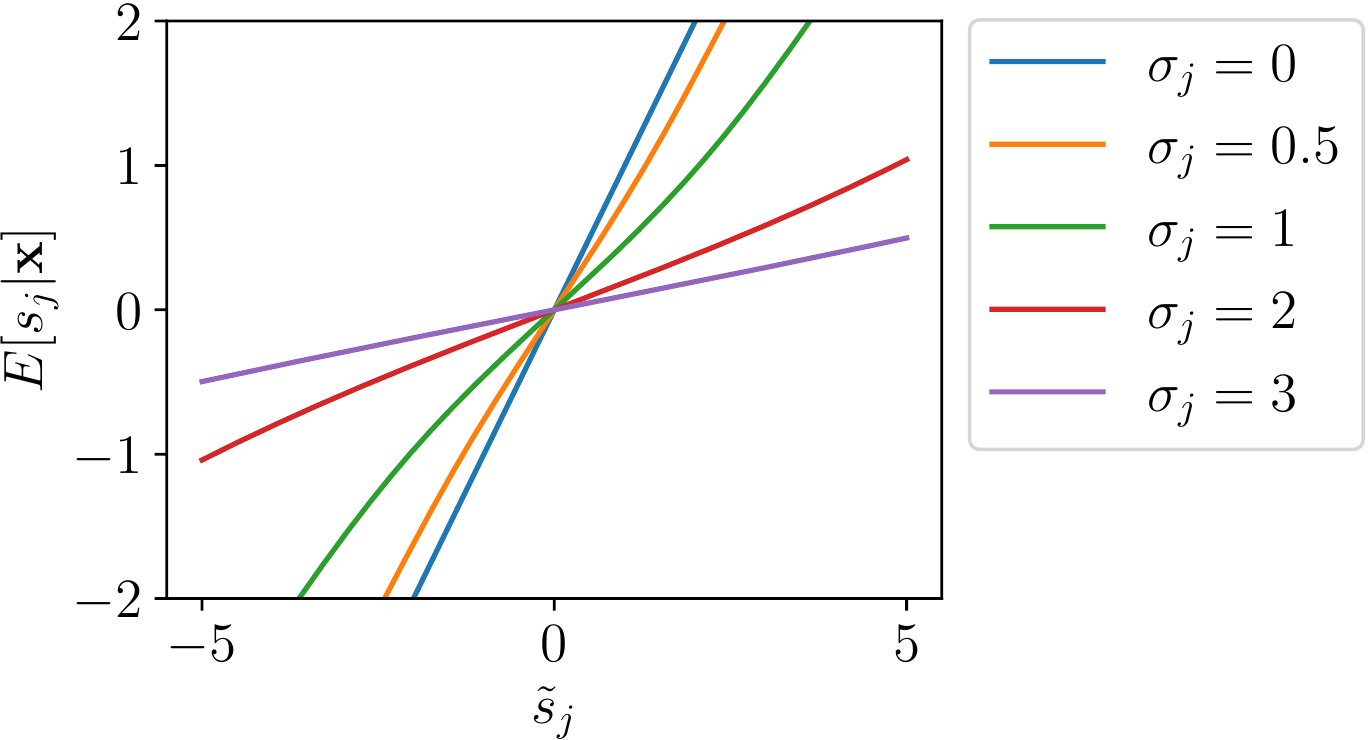}
    \caption{\textbf{MMSE estimator}: Plot of the MMSE estimator $\mathbb{E}[s_j | \xb]$ in function of $\tilde{s}_j$ and $\sigma_j$ with $m=1$.}
    \label{fig:plot_mmse}
  \end{figure}

We then obtain a MMSE estimator:
\begin{align}
  \mathbb{E}[s_j|\xb] &= \frac{\sum_{\alpha \in \{\frac12, \frac32\}}\theta_{\alpha, j} \tilde{\mu}_j}{\sum_{\alpha \in \{\frac12, \frac32\}}\theta_{\alpha, j} }
 \label{mmse}
\end{align}
which is used in the rest of the paper as our estimate of the common sources.
The MMSE estimator weights the unmixed data of each view based on the
relative noise precisions estimates and then applies a non-linear shrinkage. A
plot of $\mathbb{E}[s_j | \xb]$ in function of $\tilde{s}_j$ and $\sigma_j$ with $m=1$ is given
in Figure~\ref{fig:plot_mmse}. We see that the shrinkage is stronger for the more noisy sources.
  
\subsection{Quasi-Newton Expectation Maximization algorithm}
The second method we propose to maximize the log-likelihood is a generalized
EM~\cite{neal1998view, dempster1977maximum}, 
which we call
\emph{quasi-Newton EM} as it involves quasi-Newton steps.
The closed-form for $p(\sbb|\xb)$ makes the implementation of the E-step simple.

The complete negative log-likelihood is given by:
\begin{align*}
  &\mathcal{C} = -\log(\prod_{i=1}^m p(\xb^i|\sbb)p(\sbb)) \\
  &=  \sum_{i=1}^m \left[-\log(|W^i|) -\log(\mathcal{N}( \yb^i; \sbb, \Sigma^i))\right] + \mathrm{const}
\end{align*}
This complete negative log-likelihood is quadratic in $\sbb$ and therefore we
only need $\mathbb{E}[\sbb|\xb]$ given in~\eqref{mmse} and $\Var[\sbb|\xb]$ given by 
\begin{align*}
  \Var[s_j|\xb] &= \frac{\sum_{\alpha \in \{\frac12, \frac32\}} \theta_{\alpha, j} \tilde{\nu}_j}{\sum_{\alpha \in \{\frac12, \frac32\}}\theta_{\alpha, j} } \enspace .
\end{align*}

We then minimize the negative complete log-likelihood with respect to
$\Sigma^i$ and $W^i$.
Computing the gradient with respect to ${\Sigma^i}^{-1}$ we get the closed-form
updates for $\Sigma^i$:
\begin{align*}
&\Sigma^i \leftarrow \diag(\mathbb{E}[(\yb^i - \mathbb{E}[\sbb|\xb])(\yb^i - \mathbb{E}[\sbb|\xb])^\top) + \Var[\sbb|\xb]
\end{align*}
We update $W^i$ by performing a quasi-Newton step. The gradient $\mathcal{G}^{W^i}$
and the Hessian $\mathcal{H}^{W^i}$ are given by
\begin{align*}
  &\mathcal{G}^{W^i} = -I + (\Sigma^i)^{-1}(\yb^i - \mathbb{E}[\sbb|\xb])(\yb^i)^{\top} \\
  &\mathcal{H}^{W^i}_{a, b, c, d} = \delta_{a, c} \frac{y^i_b y^i_d}{\Sigma^i_a} \simeq \delta_{a, c} \delta_{b, d}\frac{(y^i_b)^2}{\Sigma^i_a}
\end{align*}
where we use a similar Hessian approximation as in equation~\eqref{Hessianw}.
We give detailed derivations in Appendix~\ref{gradients_em}.
Updates for $W^i$ are then given by \\
$W^i \leftarrow (I - \rho (\mathcal{H}^{W^i})^{-1} \mathcal{G}^{W^i}) W^i$, \\
where $\rho$ is chosen by backtracking line-search.
We alternate between computing the statistics $\mathbb{E}[\sbb|\xb]$ and
$\Var[\sbb|\xb]$ (E-step) and updates of parameters $\Sigma^i$ and $W^i$
(M-step). We can compute $\lambda^i_j$ and $\sigma_j$ from $(\Sigma^i_j)_{i=1}^m$ using $\sigma_j = \left(\frac1m\sum_{i=1}^m \frac1{\Sigma^i_j}\right)^{-1/2}$ and
$(\lambda^i_j)^2 =  \frac{(\sigma_j)^2}{\Sigma^i_j m}$ and monitor the convergence using the gradients with respect to the actual
log-likelihood in equation~\eqref{eq:avica_likelihood_unconstrained}.

\section{Related work}
A number of methods are available in the literature to perform group ICA.

A popular approach~\cite{calhoun2001method} is to concatenate all the views and then apply PCA in order
to obtain a reduced representation that has the same number of dimensions as in
one view. ICA is then applied on the reduced representation. We refer to this
method as \emph{ConcatICA} in the experiments.
A related approach called \emph{CanICA}~\cite{varoquaux2009canica} uses
multi-set CCA instead of PCA for fusing the views.
Lastly another fast approach is \emph{PermICA}~\cite{Esposito05NI, Hyva11NI} that
estimates the sources from each view separately. The sources are then matched
across views using the
Hungarian algorithm~\cite{tichavsky2004optimal} and the matched
sources are averaged to yield an estimate for the common sources.
These three methods are very fast but they do not optimize a proper likelihood so
they do not benefit from the advantages of such estimators such as
statistical efficiency.

The tensorial approach of~\cite{beckmann2005tensorial} imposes a particular
structure on the unmixing matrices so that unmixing matrices of different
views are row-wise scaled version of one another.
This approach can model view-specific and source-specific variability
via the row-wise scaling but imposes a structure on the unmixing matrices that may be limiting.
AVICA does not have such constraints and still enjoys identifiability.

Two extensions of~\cite{beckmann2005tensorial} are presented
in~\cite{guo2008unified}. The first extension is useful when views belong to
different groups but it keeps the same constraints on the unmixing matrices.
The second extension is more general and relaxes the constraints on the unmixing
matrices while allowing for Gaussian additive noise.
However the covariance of the additive noise is assumed to be the same for each
view. 

A more recent approach, MultiViewICA~\cite{richard2020modeling} (\emph{MVICA}) assumes noise on the
sources and proposes an efficient likelihood based approach to optimize it.
However it assumes a fixed and identical noise variance for all views and sources.
In contrast, AVICA infers the noise variance and allows it to differ
depending on views and sources. This allows AVICA to weight the estimates of the
common sources from each view based on how noisy they are, yielding in the end a better estimate of the shared sources (as will be seen in Figure~\ref{fig:syn} below).

\section{Experiments}
\label{sec:expts}

  \subsection{Dimension reduction and initialization}
  \paragraph{Dimension reduction} In this work, the number of sources is assumed to
  be equal to the number of sensors so that the mixing matrices have square shape. However, in practice the number of sensors can be much larger than the desired number of sources.
  In the experiments section we use view-specific PCA to perform dimension
  reduction when using MEG data or when performing spatial ICA on fMRI data.
  Following the suggestion of~\cite{richard2020modeling} we use the shared
  response model (SRM,~\cite{chen2015reduced}) to
  reduce the data dimension when applying temporal ICA on fMRI data.
  Note that the choice of dimension reduction technique generally has an impact on the results. However we leave this discussion to future work.

  \paragraph{Initialization} The ICA problem is non convex, therefore the result depends on its
  initialization. ConcatICA, PermICA and CanICA are randomly initialized. AVICA and MVICA
  are initialized using ConcatICA.

\paragraph{Software tools}
Experiments used Nilearn~\cite{abraham2014machine} and MNE~\cite{gramfort2013meg} for fMRI and MEG data
processing respectively, as well as the scientific Python ecosystem:
Matplotlib~\cite{hunter2007matplotlib}, Scikit-learn~\cite{pedregosa2011scikit},
Numpy~\cite{harris2020array}, Scipy~\cite{2020SciPy-NMeth} and Sympy~\cite{sympy}.

\subsection{Synthetic experiments}
\label{synth_exp}
We first validate the proposed estimator for the AVICA model using synthetic data generated according to the model in equation~\eqref{eq:avica}.
The sources are generated i.i.d. from a Laplace density
$\delta(x)=\frac1{\sqrt{2}}\exp(-\sqrt{2}|x|)$.

\paragraph{Convergence plot}
We use $m=10$ views and $k=5$ sources. Mixing matrices $(A^i)_{i=1}^m$ are generated with i.i.d. entries following a normal distribution.
For each source, the relative noise precisions $(({\lambda^i_j})^2)_{i=1}^m$ are
generated from a Dirichlet distribution with parameter $(1 \dots 1)$.
The log of the global noise levels $(\log(\sigma_j))_{j=1}^k$
are generated from a normal distribution where the mean is fixed to 0 and the variance is fixed to $\frac12$.
We generate $n=1000$ samples.
We set $\mu=0$ so that the two methods
optimize the exact same log-likelihood.

\begin{figure}
  \centering{
    \includegraphics[width=0.55\textwidth]{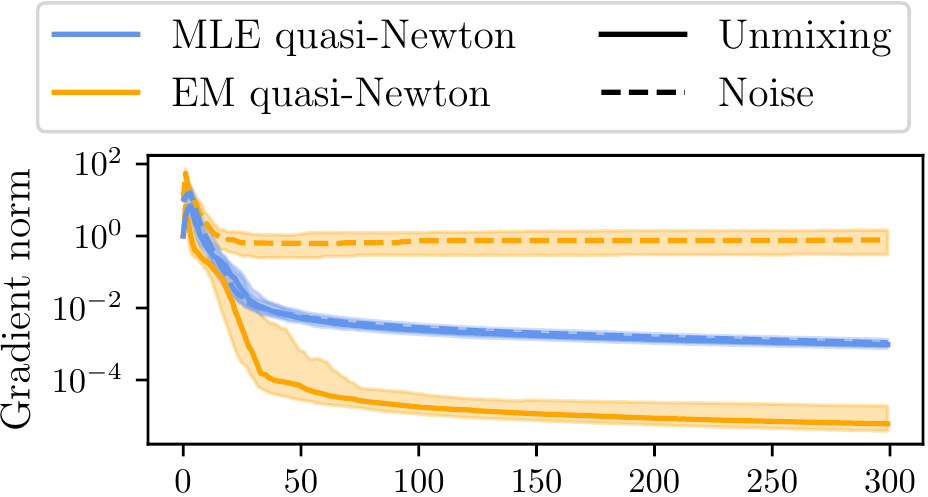}}
  \centering{
  \includegraphics[width=0.55\textwidth]{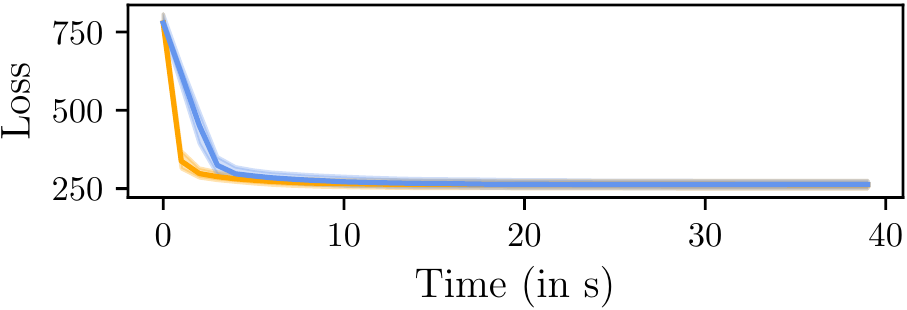}
  }
  \caption{\textbf{Synthetic experiment - Convergence plot}: Median gradient
    norm (top) and negative log-likelihood (bottom) as a function of time. Dashed lines correspond to the gradient with
    respect to noise parameters and solid lines to the gradient with respect to
    unmixing matrices. Error bars display the first and last quartiles.}
  \label{fig:em_vs_gd}
\end{figure}

For each optimizer we run the following analysis 100 times with 100 different seeds.
At each iteration, we record the $\ell_\infty$ norm of the gradient with respect
to the noise parameters, the $\ell_\infty$ norm of the gradient with respect to
the unmixing matrices, the negative log-likelihood and the current time.
We interpolate between time points so that we can present convergence curves in
function of computation time in Figure~\ref{fig:em_vs_gd}.
 We report the median value, and error bars correspond to the first
and last quartiles. 

The EM quasi-Newton is slightly faster. However the homogeneous decrease of the
norm of the different gradients in MLE quasi-Newton makes it easier to
monitor. In our experiments we use MLE quasi-Newton as our optimizer. Other
convergence plots are available in appendix~\ref{app:other_convergence_plot}

\paragraph{Improved reconstruction error of AVICA}
We use the same generative model as above but the log of the global noise level
$(\log(\sigma_j))_{j=1}^k$ are generated from a normal distribution
where the mean varies between $-2$ and $2$ (the variance is still fixed to $\frac12$).
Each compared algorithm returns an estimates of the sources.
The performance is measured by computing the reconstruction
error which is defined as $1 - \mathbb{E}[\sbb \hat{\sbb}]$ where $\sbb$ are the true
sources and $\hat{\sbb}$ are the estimated sources both normalized so that they
have unit variance.
Each experiment is repeated $100$ times with $100$ different seeds.  We report the median value, and error bars correspond to the first
and last quartiles. 

\begin{figure}
  \centering
  \includegraphics[width=0.55\textwidth]{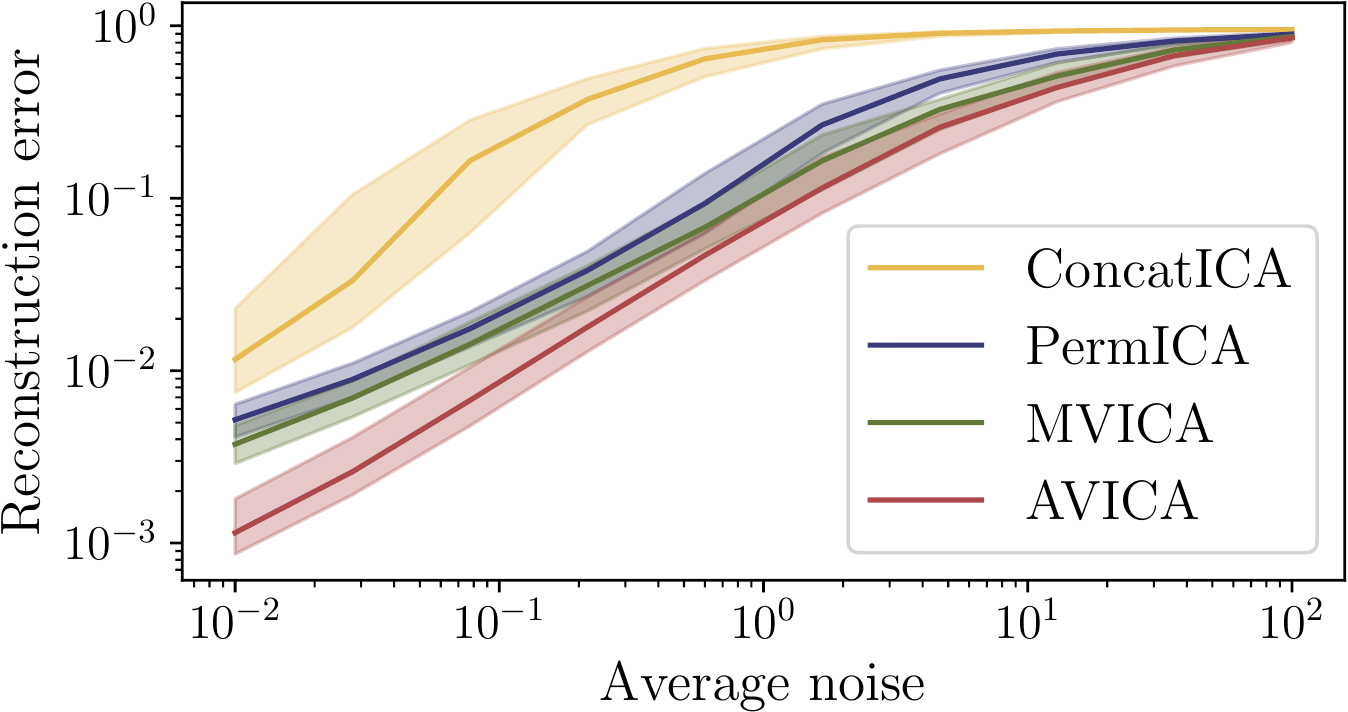}
  \caption{\textbf{Synthetic experiment - Reconstruction error}: Median distance
    between true and recovered sources. Error bars display the first and last quartiles.}
  \label{fig:syn}
\end{figure}

Figure~\ref{fig:syn} shows that Adaptive multiViewICA outperforms other
approaches even though the density used to generate the data is not the same as
the one used in the model.

Additional experiments show in appendix~\ref{paramident} that
AVICA recovers well the relative noise precisions parameters and in
appendix~\ref{adaptivescaling} that AVICA prioritizes informative views.

\subsection{fMRI experiments}
\label{sec:fmri_exp}

\textbf{fMRI data and preprocessing} 
We use four different fMRI datasets. For each dataset, we specify the number
of views $m$ which is the number of subjects and the number
of samples $n$ which is the number of brain images per subject. Due to the acquisition pipeline, the samples are generally split into
groups of approximately identical size called \emph{runs}.
The data have been preprocessed following standard procedures for fMRI data, including realignment within and across runs, and resampling to the template MNI space. 
Data have been band-pass filtered in the temporal domain in the $[.01, .1]$Hz range.
The \emph{sherlock}
dataset~\cite{chen2017shared} contains $m=16$ subjects and $n=1976$ brain images
per subject split into $5$
runs. The subjects are watching one
episode of the ``Sherlock'' TV show.
The \emph{forrest} dataset~\cite{hanke2014high} contains $m=16$ subjects and $n=3261$ brain images per subject split into $7$ runs.
The subjects are listening to an audio version of the ``Forrest Gump'' movie.
The \emph{raiders} dataset~\cite{ibc} contains $m=11$ subjects and $n=3252$ brain images
per subject split into $10$ runs.
The subjects are watching the ``Raiders of the Lost ark'' movie.
The \emph{clips} dataset~\cite{ibc} contains $m=12$ subjects and $n=3900$ brain images
per subject split into $12$ runs. The subjects are watching video clips (without audio).

\textbf{Reconstructing the BOLD signal of missing subjects}
We reproduce the experimental pipeline of~\cite{richard2020modeling} to benchmark
group ICA methods using their ability to reconstruct fMRI data of a left-out
subject.

All data  undergo a 6\,mm spatial smoothing and we focus on the same regions of interests
(ROI) as in~\cite{richard2020modeling} that are also reported in
Appendix~\ref{app:roi}.
We call a \emph{forward operator} the product of the dimension
reduction operator and an unmixing matrix and a \emph{backward operator} its pseudo
inverse. There is one forward operator and one backward operator per view.

The forward operators are learned using all subjects
and $80\%$ of the runs. Then they are applied on the remaining $20\%$ of the runs using $80\%$
of the subjects yielding unmixed data. In order to obtain an estimate of the
shared sources the unmixed data are averaged. When AVICA is used, this average
becomes a weighted average where the weights are given by $(\lambda^i_j)^2$. We
then apply the backward operator of the remaining $20\%$ subjects on the shared sources to
reconstruct their data.

The accuracy of the reconstruction is measured via the $R^2$ score which
measures for each voxel the discrepancy between the true timecourse $\zb$ and the
predicted timecourse $\hat{\zb}$:
$R^2(\hat{\zb}, \zb) = 1 - \frac{\sum_{t=1}^n (z_t - \hat{z_t})^2}{\sum_{t=1}^n
  (z_t - \bar{z})^2}$
where $\bar{z} = \frac1n \sum_{t=1}^n z_t$ is the empirical mean of $\zb$.

\begin{figure}
  \centering
  \includegraphics[width=0.55\textwidth]{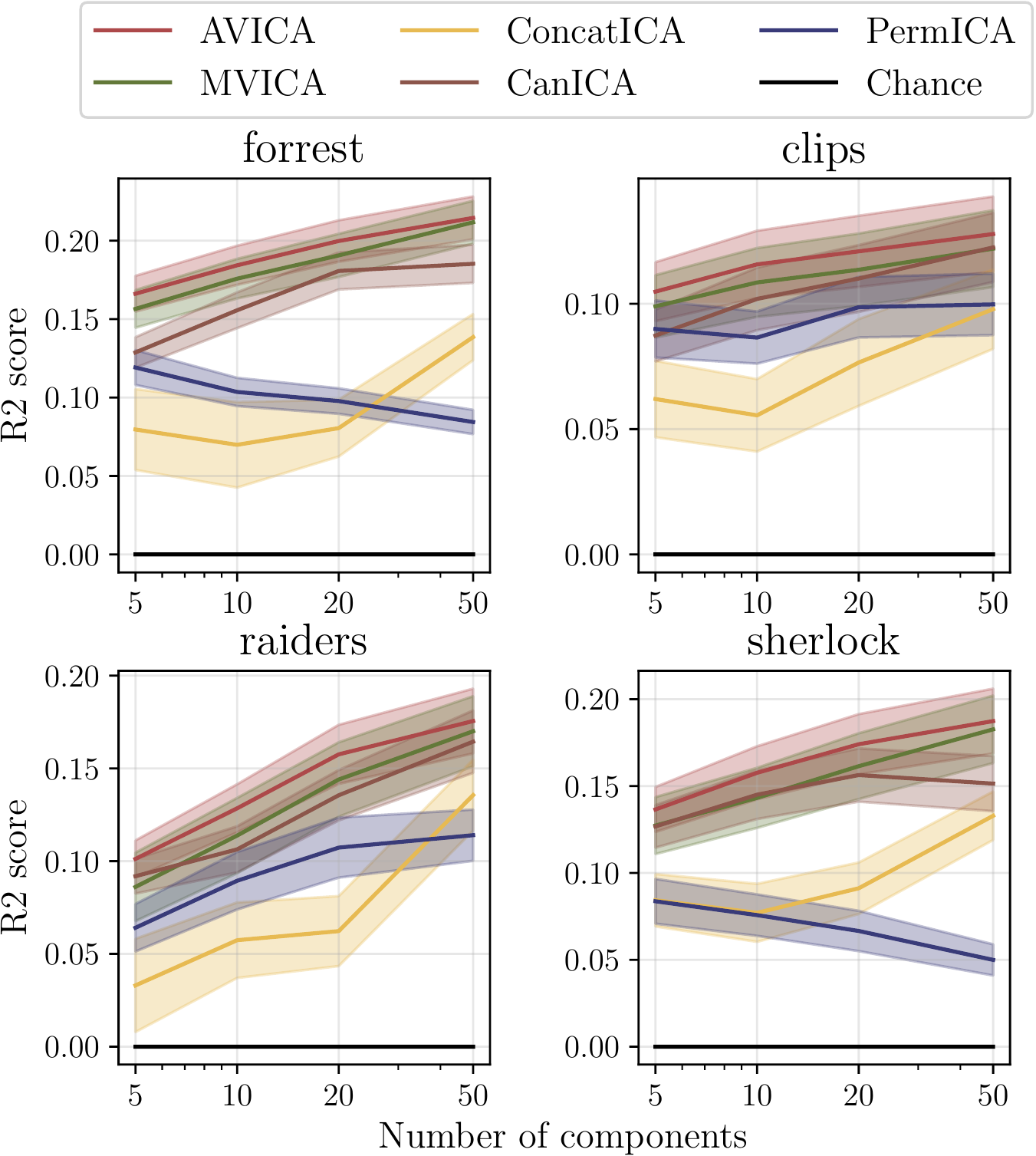}
  \caption{\textbf{Reconstructing the BOLD signal of
      missing subjects}. Mean $R^2$ score between reconstructed data and true
    data (higher is better). Error bars represent a 95 \% confidence interval.}
  \label{fig:reconstruction}
\end{figure}

For each compared algorithm, the experiment is run 25 times with different seeds
to obtain error bars. We report the mean $R^2$ score across voxels
 and display the
results in Figure~\ref{fig:reconstruction}. The error bars represent a $95\%$
confidence interval.
The chance level is given by the performance of an algorithm that computes its
dimension reduction operator and unmixing matrices by sampling from a standard
normal distribution.

AVICA exhibits a small improvement on sherlock and raiders datasets that is
consistent for different numbers of components (i.e. numbers of sources) and competitive results on clips and forrest datasets.

In appendix~\ref{fmristability}, an experiment on rest fMRI data
shows that AVICA yields more stable sources than other approaches.

\subsection{MEG experiments}
In the following experiments we consider the Cam-CAN
dataset~\cite{taylor2017cambridge}. We use the magnetometer data from the MEG of $m=496$ subjects.
Each subject is repeatedly presented three audio-visual stimuli. 
For each stimulus, we divide the trials into two sets and within each set, 
the MEG signal is averaged across trials to isolate the evoked response. This
procedure yields 6 chunks of individual data.

The 6 chunks of data are concatenated in the time direction and ICA algorithms
are applied separately to extract $k=10$ shared sources that we plot in
appendix~\ref{megsources} and localize in appendix~\ref{megsourceslocal}.

\paragraph{Robustness w.r.t intra-subject variability in MEG}
We first study the similarity between group sources corresponding to repetitions of the same
stimuli. This gives a measure of robustness of each ICA algorithm with respect
to intra-subject variability, or sampling noise.

\begin{figure}
  \centering
  \includegraphics[width=0.55
  \textwidth]{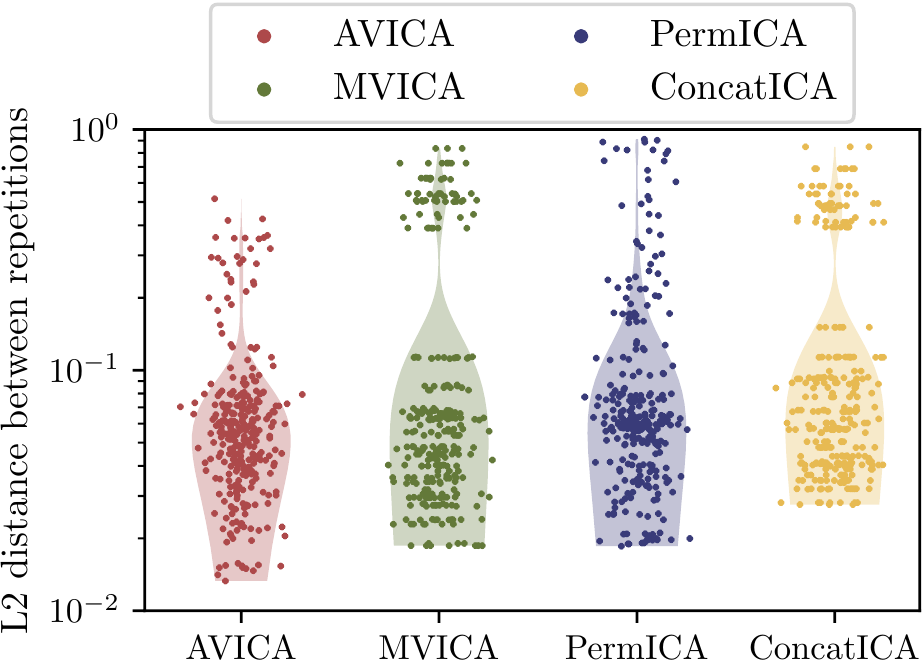}
  \caption{\textbf{Robustness w.r.t intra-subject variability in MEG}:
    Algorithms are run 10 times with different seeds on preprocessed MEG CamCAN data
    containing 6 chunks of data (2 repetitions for each of the 3 stimulus). We
    display a scatter plot showing for each algorithm the distance between group
    sources corresponding to repetitions of the same
    stimulus. 
}
\label{fig:eeg_intragroup_variability}
\end{figure}

Algorithms are run 10 times with different seeds on the 6 chunks of data,
and group sources are extracted.
When two chunks of data correspond to repetitions of the same stimulus they should yield similar
sources.
For each source and for each stimulus, we therefore measure the $\ell_2$
distance between the two repetitions of the stimulus.
 This yields $300$ distances per algorithm that are
plotted on Figure~\ref{fig:eeg_intragroup_variability}.

The sources recovered by AVICA have a lower variability than other approaches.
The difference between AVICA and other approaches can be quantified via a
statistical t-test on the difference of log distances. It
gives the p-values $4.57 \times 10^{-4}$, $5.93 \times 10^{-4}$ and $2.37
\times 10^{-9}$ w.r.t.
MVICA, PermICA and ConcatICA respectively.

\paragraph{Recovering the correct noise level in MEG data}
Here, we study the ability of AVICA to correctly estimate the
noise level. For each subject we concatenate the signal corresponding to the 6
chunks of data and compute the mean square error of the baseline 
(spontaneous) activity
- when the
subject is not exposed to any stimulus. This quantity is assumed to reflect the
true noise level of this subject which we compare to the noise level estimated
by AVICA: $\tr(A_i \Sigma_i A_i^{\top})$ where $A_i = W_i^{-1}$.

\begin{figure}
  \centering
  \includegraphics[width=0.55 \textwidth]{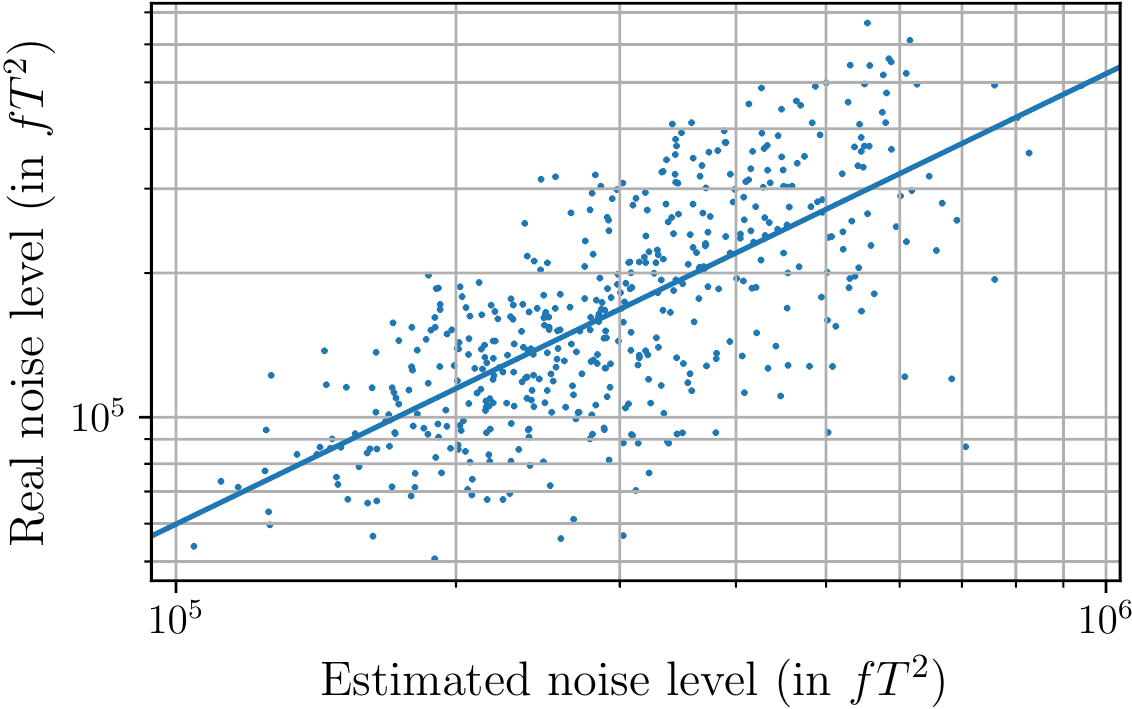}
  \caption{\textbf{Estimating noise level in  CamCAN MEG}: Scatter plot showing the true noise level and
    the noise level estimated by Adaptive multiViewICA. We fit a linear
    regression on the log of the true and predicted noise level: $r^2=0.50$ ($p<10^{-74}$).}
  \label{fig:eeg_scatter}
\end{figure}

Figure~\ref{fig:eeg_scatter} shows a scatter plot that compares the estimated
and true noise level. We take the log of the predicted and true noise level and
fit a linear regression: we obtain a $r^2$ score of $0.50$ and
a p-value of the slope below $10^{-74}$.

An additional MEG experiments on phantom data in appendix~\ref{megphantom} demonstrates that
AVICA recovers better source estimates than other approaches.

\section*{Conclusion}
\label{sec:disc}
We introduced AVICA, a principled method to solve the group ICA problem, which is an important case of multiview learning. The model explicitly estimates noise levels for each subject and source, which has not hitherto been possible.  
AVICA enables a closed-form likelihood, which can be minimized quickly with an alternate quasi-Newton method or a generalized EM algorithm. Unlike previous models, AVICA also enables a closed-form MMSE estimator of the sources.

AVICA exhibits performance superior to other group ICA methods on synthetic data and real data involving two different neuroimaging modalities.
Experiments on neuroimaging data suggest that AVICA is the method of choice to fuse data across
subjects. Future work could investigate whether the common representation
learned by AVICA leads to a plausible template of the human brain, as well as the interplay with dimension reduction.

\bibliography{biblio_amvica}

\begin{thebibliography}{38}
\providecommand{\natexlab}[1]{#1}
\providecommand{\url}[1]{\texttt{#1}}
\expandafter\ifx\csname urlstyle\endcsname\relax
  \providecommand{\doi}[1]{doi: #1}\else
  \providecommand{\doi}{doi: \begingroup \urlstyle{rm}\Url}\fi

\bibitem[Ablin et~al.(2018)Ablin, Cardoso, and Gramfort]{ablin2018faster}
Ablin, P., Cardoso, J.-F., and Gramfort, A.
\newblock Faster independent component analysis by preconditioning with
  {H}essian approximations.
\newblock \emph{IEEE Transactions on Signal Processing}, 66\penalty0
  (15):\penalty0 4040--4049, 2018.

\bibitem[Abraham et~al.(2014)Abraham, Pedregosa, Eickenberg, Gervais, Mueller,
  Kossaifi, Gramfort, Thirion, and Varoquaux]{abraham2014machine}
Abraham, A., Pedregosa, F., Eickenberg, M., Gervais, P., Mueller, A., Kossaifi,
  J., Gramfort, A., Thirion, B., and Varoquaux, G.
\newblock Machine learning for neuroimaging with scikit-learn.
\newblock \emph{Frontiers in neuroinformatics}, 8:\penalty0 14, 2014.

\bibitem[Absil \& Malick(2012)Absil and Malick]{absil2012projection}
Absil, P.-A. and Malick, J.
\newblock Projection-like retractions on matrix manifolds.
\newblock \emph{SIAM Journal on Optimization}, 22\penalty0 (1):\penalty0
  135--158, 2012.

\bibitem[Beckmann \& Smith(2005)Beckmann and Smith]{beckmann2005tensorial}
Beckmann, C.~F. and Smith, S.~M.
\newblock Tensorial extensions of independent component analysis for
  multisubject {fMRI} analysis.
\newblock \emph{Neuroimage}, 25\penalty0 (1):\penalty0 294--311, 2005.

\bibitem[Bell \& Sejnowski(1995)Bell and Sejnowski]{bell1995information}
Bell, A.~J. and Sejnowski, T.~J.
\newblock An information-maximization approach to blind separation and blind
  deconvolution.
\newblock \emph{Neural computation}, 7\penalty0 (6):\penalty0 1129--1159, 1995.

\bibitem[Calhoun et~al.(2001)Calhoun, Adali, Pearlson, and
  Pekar]{calhoun2001method}
Calhoun, V.~D., Adali, T., Pearlson, G.~D., and Pekar, J.~J.
\newblock A method for making group inferences from functional {MRI} data using
  independent component analysis.
\newblock \emph{Human brain mapping}, 14\penalty0 (3):\penalty0 140--151, 2001.

\bibitem[Cardoso(1997)]{cardoso1997infomax}
Cardoso, J.-F.
\newblock Infomax and maximum likelihood for blind source separation.
\newblock \emph{IEEE Signal processing letters}, 4\penalty0 (4):\penalty0
  112--114, 1997.

\bibitem[Chen et~al.(2017)Chen, Leong, Honey, Yong, Norman, and
  Hasson]{chen2017shared}
Chen, J., Leong, Y.~C., Honey, C.~J., Yong, C.~H., Norman, K.~A., and Hasson,
  U.
\newblock Shared memories reveal shared structure in neural activity across
  individuals.
\newblock \emph{Nature neuroscience}, 20\penalty0 (1):\penalty0 115--125, 2017.

\bibitem[Chen et~al.(2015)Chen, Chen, Yeshurun, Hasson, Haxby, and
  Ramadge]{chen2015reduced}
Chen, P.-H., Chen, J., Yeshurun, Y., Hasson, U., Haxby, J., and Ramadge, P.~J.
\newblock A reduced-dimension {fMRI} shared response model.
\newblock In \emph{Advances in Neural Information Processing Systems}, pp.\
  460--468, 2015.

\bibitem[Chen et~al.(2007)Chen, H{\"a}rdle, and Spokoiny]{chen2007portfolio}
Chen, Y., H{\"a}rdle, W., and Spokoiny, V.
\newblock Portfolio value at risk based on independent component analysis.
\newblock \emph{Journal of Computational and Applied Mathematics}, 205\penalty0
  (1):\penalty0 594--607, 2007.

\bibitem[Comon(1994)]{comon1994independent}
Comon, P.
\newblock Independent component analysis, a new concept?
\newblock \emph{Signal processing}, 36\penalty0 (3):\penalty0 287--314, 1994.

\bibitem[Dempster et~al.(1977)Dempster, Laird, and Rubin]{dempster1977maximum}
Dempster, A.~P., Laird, N.~M., and Rubin, D.~B.
\newblock Maximum likelihood from incomplete data via the em algorithm.
\newblock \emph{Journal of the Royal Statistical Society: Series B
  (Methodological)}, 39\penalty0 (1):\penalty0 1--22, 1977.

\bibitem[Esposito et~al.(2005)Esposito, Scarabino, Hyv\"arinen, Himberg,
  Formisano, Comani, Tedeschi, Goebel, Seifritz, and Salle]{Esposito05NI}
Esposito, F., Scarabino, T., Hyv\"arinen, A., Himberg, J., Formisano, E.,
  Comani, S., Tedeschi, G., Goebel, R., Seifritz, E., and Salle, F.~D.
\newblock Independent component analysis of {fMRI} group studies by
  self-organizing clustering.
\newblock \emph{NeuroImage}, 25\penalty0 (1):\penalty0 193--205, 2005.

\bibitem[Gramfort et~al.(2013)Gramfort, Luessi, Larson, Engemann, Strohmeier,
  Brodbeck, Goj, Jas, Brooks, Parkkonen, et~al.]{gramfort2013meg}
Gramfort, A., Luessi, M., Larson, E., Engemann, D.~A., Strohmeier, D.,
  Brodbeck, C., Goj, R., Jas, M., Brooks, T., Parkkonen, L., et~al.
\newblock {MEG} and {EEG} data analysis with {MNE-Python}.
\newblock \emph{Frontiers in neuroscience}, 7:\penalty0 267, 2013.

\bibitem[Guo \& Pagnoni(2008)Guo and Pagnoni]{guo2008unified}
Guo, Y. and Pagnoni, G.
\newblock A unified framework for group independent component analysis for
  multi-subject {fMRI} data.
\newblock \emph{NeuroImage}, 42\penalty0 (3):\penalty0 1078--1093, 2008.

\bibitem[Hanke et~al.(2014)Hanke, Baumgartner, Ibe, Kaule, Pollmann, Speck,
  Zinke, and Stadler]{hanke2014high}
Hanke, M., Baumgartner, F.~J., Ibe, P., Kaule, F.~R., Pollmann, S., Speck, O.,
  Zinke, W., and Stadler, J.
\newblock A high-resolution {7-Tesla} {fMRI} dataset from complex natural
  stimulation with an audio movie.
\newblock \emph{Scientific data}, 1:\penalty0 140003, 2014.

\bibitem[Harris et~al.(2020)Harris, Millman, van~der Walt, Gommers, Virtanen,
  Cournapeau, Wieser, Taylor, Berg, Smith, Kern, Picus, Hoyer, van Kerkwijk,
  Brett, Haldane, del R{'{\i}}o, Wiebe, Peterson, G{'{e}}rard-Marchant,
  Sheppard, Reddy, Weckesser, Abbasi, Gohlke, and Oliphant]{harris2020array}
Harris, C.~R., Millman, K.~J., van~der Walt, S.~J., Gommers, R., Virtanen, P.,
  Cournapeau, D., Wieser, E., Taylor, J., Berg, S., Smith, N.~J., Kern, R.,
  Picus, M., Hoyer, S., van Kerkwijk, M.~H., Brett, M., Haldane, A., del
  R{'{\i}}o, J.~F., Wiebe, M., Peterson, P., G{'{e}}rard-Marchant, P.,
  Sheppard, K., Reddy, T., Weckesser, W., Abbasi, H., Gohlke, C., and Oliphant,
  T.~E.
\newblock Array programming with {NumPy}.
\newblock \emph{Nature}, 585\penalty0 (7825):\penalty0 357--362, September
  2020.
\newblock \doi{10.1038/s41586-020-2649-2}.
\newblock URL \url{https://doi.org/10.1038/s41586-020-2649-2}.

\bibitem[Hunter(2007)]{hunter2007matplotlib}
Hunter, J.~D.
\newblock Matplotlib: A 2d graphics environment.
\newblock \emph{Computing in science \& engineering}, 9\penalty0 (3):\penalty0
  90--95, 2007.

\bibitem[Hyv{\"a}rinen(1998)]{hyvarinen1998independent}
Hyv{\"a}rinen, A.
\newblock Independent component analysis in the presence of gaussian noise by
  maximizing joint likelihood.
\newblock \emph{Neurocomputing}, 22\penalty0 (1-3):\penalty0 49--67, 1998.

\bibitem[Hyv{\"a}rinen(2011)]{Hyva11NI}
Hyv{\"a}rinen, A.
\newblock Testing the {ICA} mixing matrix based on inter-subject or
  inter-session consistency.
\newblock \emph{NeuroImage}, 58\penalty0 (1):\penalty0 122--136, 2011.

\bibitem[Hyv{\"a}rinen \& Oja(2000)Hyv{\"a}rinen and
  Oja]{hyvarinen2000independent}
Hyv{\"a}rinen, A. and Oja, E.
\newblock Independent component analysis: algorithms and applications.
\newblock \emph{Neural networks}, 13\penalty0 (4-5):\penalty0 411--430, 2000.

\bibitem[Liebermeister(2002)]{liebermeister2002linear}
Liebermeister, W.
\newblock Linear modes of gene expression determined by independent component
  analysis.
\newblock \emph{Bioinformatics}, 18\penalty0 (1):\penalty0 51--60, 2002.

\bibitem[Liu(2016)]{liu2016noise}
Liu, T.~T.
\newblock Noise contributions to the fmri signal: An overview.
\newblock \emph{NeuroImage}, 143:\penalty0 141--151, 2016.

\bibitem[Maino et~al.(2002)Maino, Farusi, Baccigalupi, Perrotta, Banday,
  Bedini, Burigana, De~Zotti, G{\'o}rski, and Salerno]{maino2002all}
Maino, D., Farusi, A., Baccigalupi, C., Perrotta, F., Banday, A., Bedini, L.,
  Burigana, C., De~Zotti, G., G{\'o}rski, K., and Salerno, E.
\newblock All-sky astrophysical component separation with fast independent
  component analysis (fastica).
\newblock \emph{Monthly Notices of the Royal Astronomical Society},
  334\penalty0 (1):\penalty0 53--68, 2002.

\bibitem[Meurer et~al.(2017)Meurer, Smith, Paprocki, \v{C}ert\'{i}k, Kirpichev,
  Rocklin, Kumar, Ivanov, Moore, Singh, Rathnayake, Vig, Granger, Muller,
  Bonazzi, Gupta, Vats, Johansson, Pedregosa, Curry, Terrel, Rou\v{c}ka, Saboo,
  Fernando, Kulal, Cimrman, and Scopatz]{sympy}
Meurer, A., Smith, C.~P., Paprocki, M., \v{C}ert\'{i}k, O., Kirpichev, S.~B.,
  Rocklin, M., Kumar, A., Ivanov, S., Moore, J.~K., Singh, S., Rathnayake, T.,
  Vig, S., Granger, B.~E., Muller, R.~P., Bonazzi, F., Gupta, H., Vats, S.,
  Johansson, F., Pedregosa, F., Curry, M.~J., Terrel, A.~R., Rou\v{c}ka, v.,
  Saboo, A., Fernando, I., Kulal, S., Cimrman, R., and Scopatz, A.
\newblock Sympy: symbolic computing in python.
\newblock \emph{PeerJ Computer Science}, 3:\penalty0 e103, January 2017.
\newblock ISSN 2376-5992.
\newblock \doi{10.7717/peerj-cs.103}.
\newblock URL \url{https://doi.org/10.7717/peerj-cs.103}.

\bibitem[Neal \& Hinton(1998)Neal and Hinton]{neal1998view}
Neal, R.~M. and Hinton, G.~E.
\newblock A view of the em algorithm that justifies incremental, sparse, and
  other variants.
\newblock In \emph{Learning in graphical models}, pp.\  355--368. Springer,
  1998.

\bibitem[Nocedal \& Wright(2006)Nocedal and Wright]{nocedal2006numerical}
Nocedal, J. and Wright, S.
\newblock \emph{Numerical optimization}.
\newblock Springer Science \& Business Media, 2006.

\bibitem[Pascual-Marqui et~al.(2002)]{pascual2002standardized}
Pascual-Marqui, R.~D. et~al.
\newblock Standardized low-resolution brain electromagnetic tomography
  ({sLORETA}): technical details.
\newblock \emph{Methods Find Exp Clin Pharmacol}, 24\penalty0 (Suppl
  D):\penalty0 5--12, 2002.

\bibitem[Pedregosa et~al.(2011)Pedregosa, Varoquaux, Gramfort, Michel, Thirion,
  Grisel, Blondel, Prettenhofer, Weiss, Dubourg, et~al.]{pedregosa2011scikit}
Pedregosa, F., Varoquaux, G., Gramfort, A., Michel, V., Thirion, B., Grisel,
  O., Blondel, M., Prettenhofer, P., Weiss, R., Dubourg, V., et~al.
\newblock Scikit-learn: {Machine} learning in {Python}.
\newblock \emph{the Journal of machine Learning research}, 12:\penalty0
  2825--2830, 2011.

\bibitem[Penny \& Holmes(2007)Penny and Holmes]{penny2007random}
Penny, W. and Holmes, A.
\newblock Random effects analysis.
\newblock \emph{Statistical parametric mapping: The analysis of functional
  brain images}, 156:\penalty0 165, 2007.

\bibitem[Pinho et~al.(2018)Pinho, Amadon, Ruest, Fabre, Dohmatob, Denghien,
  Ginisty, Becuwe-Desmidt, Roger, Laurier, et~al.]{ibc}
Pinho, A.~L., Amadon, A., Ruest, T., Fabre, M., Dohmatob, E., Denghien, I.,
  Ginisty, C., Becuwe-Desmidt, S., Roger, S., Laurier, L., et~al.
\newblock Individual brain charting, a high-resolution {fMRI} dataset for
  cognitive mapping.
\newblock \emph{Scientific data}, 5, 2018.

\bibitem[Richard et~al.(2020)Richard, Gresele, Hyvarinen, Thirion, Gramfort,
  and Ablin]{richard2020modeling}
Richard, H., Gresele, L., Hyvarinen, A., Thirion, B., Gramfort, A., and Ablin,
  P.
\newblock Modeling shared responses in neuroimaging studies through multiview
  ica.
\newblock In \emph{Advances in Neural Information Processing Systems 33},
  December 2020.

\bibitem[Ristaniemi(1999)]{ristaniemi1999performance}
Ristaniemi, T.
\newblock On the performance of blind source separation in cdma downlink.
\newblock In \emph{Proceedings of the International Workshop on Independent
  Component Analysis and Signal Separation (ICA'99)}, pp.\  437--441, 1999.

\bibitem[Taylor et~al.(2017)Taylor, Williams, Cusack, Auer, Shafto, Dixon,
  Tyler, Henson, et~al.]{taylor2017cambridge}
Taylor, J.~R., Williams, N., Cusack, R., Auer, T., Shafto, M.~A., Dixon, M.,
  Tyler, L.~K., Henson, R.~N., et~al.
\newblock The {Cambridge Centre for Ageing and Neuroscience} ({Cam-CAN}) data
  repository: structural and functional {MRI}, {MEG}, and cognitive data from a
  cross-sectional adult lifespan sample.
\newblock \emph{Neuroimage}, 144:\penalty0 262--269, 2017.

\bibitem[Tichavsky \& Koldovsky(2004)Tichavsky and
  Koldovsky]{tichavsky2004optimal}
Tichavsky, P. and Koldovsky, Z.
\newblock Optimal pairing of signal components separated by blind techniques.
\newblock \emph{IEEE Signal Processing Letters}, 11\penalty0 (2):\penalty0
  119--122, 2004.

\bibitem[Van~Essen et~al.(2013)Van~Essen, Smith, Barch, Behrens, Yacoub,
  Ugurbil, Consortium, et~al.]{van2013wu}
Van~Essen, D.~C., Smith, S.~M., Barch, D.~M., Behrens, T.~E., Yacoub, E.,
  Ugurbil, K., Consortium, W.-M.~H., et~al.
\newblock The {WU-Minn} human connectome project: an overview.
\newblock \emph{Neuroimage}, 80:\penalty0 62--79, 2013.

\bibitem[Varoquaux et~al.(2009)Varoquaux, Sadaghiani, Poline, and
  Thirion]{varoquaux2009canica}
Varoquaux, G., Sadaghiani, S., Poline, J.-B., and Thirion, B.
\newblock Can{ICA}: Model-based extraction of reproducible group-level ica
  patterns from {fMRI} time series.
\newblock \emph{arXiv preprint arXiv:0911.4650}, 2009.

\bibitem[Virtanen et~al.(2020)Virtanen, Gommers, Oliphant, Haberland, Reddy,
  Cournapeau, Burovski, Peterson, Weckesser, Bright, {van der Walt}, Brett,
  Wilson, Millman, Mayorov, Nelson, Jones, Kern, Larson, Carey, Polat, Feng,
  Moore, {VanderPlas}, Laxalde, Perktold, Cimrman, Henriksen, Quintero, Harris,
  Archibald, Ribeiro, Pedregosa, {van Mulbregt}, and {SciPy 1.0
  Contributors}]{2020SciPy-NMeth}
Virtanen, P., Gommers, R., Oliphant, T.~E., Haberland, M., Reddy, T.,
  Cournapeau, D., Burovski, E., Peterson, P., Weckesser, W., Bright, J., {van
  der Walt}, S.~J., Brett, M., Wilson, J., Millman, K.~J., Mayorov, N., Nelson,
  A. R.~J., Jones, E., Kern, R., Larson, E., Carey, C.~J., Polat, {\.I}., Feng,
  Y., Moore, E.~W., {VanderPlas}, J., Laxalde, D., Perktold, J., Cimrman, R.,
  Henriksen, I., Quintero, E.~A., Harris, C.~R., Archibald, A.~M., Ribeiro,
  A.~H., Pedregosa, F., {van Mulbregt}, P., and {SciPy 1.0 Contributors}.
\newblock {{SciPy} 1.0: Fundamental Algorithms for Scientific Computing in
  Python}.
\newblock \emph{Nature Methods}, 17:\penalty0 261--272, 2020.
\newblock \doi{10.1038/s41592-019-0686-2}.

\end{thebibliography}

\newpage
\appendix
\onecolumn

\section{Likelihood}

\subsection{The closed-form likelihood of AVICA}
\label{appendix:integration}
Here show the full derivation for likelihood computation
\begin{align}
  \loss &= -\log\left(\int_{\sbb} p(\xb|\sbb) p(\sbb) d\sbb \right) \\
  &= -\log\left(\int_{\sbb} \prod_i p(\xb^i|\sbb) p(\sbb) d\sbb \right) \label{12} \\
  &= -\log\left(\int_{\sbb} \prod_i \left[|W^i|\frac{\exp(-\sum_{j=1}^k \frac{m (\lambda^i_j)^2}{2(\sigma_j)^2}(y^i_j - s_j)^2)}{\prod_{j=1}^k  \sqrt{2\pi \frac{(\sigma_j)^2}{m (\lambda^i_j)^2}}} \right]p(\sbb) d\sbb \right) \label{13}\\
  &= \sum_{i=1}^m \left[-\log(|W^i|) +\frac1{2}\sum_{j=1}^k \log \left(\frac{(\sigma_j)^2}{({\lambda^i_j})^2 m}\right)\right] -\log (\mathcal{J}) + \mathrm{const} \label{14} 
\end{align}
with
\begin{align*}
\mathcal{J} =  \int_{\sbb}\exp\left(-\sum_{i=1}^m
  \sum_{j=1}^k\frac{({\lambda^i_j})^2m}{2(\sigma_j)^2} (y^i_j -
  s_j)^2\right)p(\sbb) d\sbb
\end{align*}
 
In~\eqref{12} we use the conditional independence of $\xb^i$ given $\sbb$,
in~\eqref{13} we make the change of variable $\nb^i = \yb^i - \sbb$ where $\yb^i
= W^i \xb^i$ and use the Gaussian assumption on $\nb^i$. We obtain~\eqref{14}
which is identical to~\eqref{eq:likelihood} up to a constant. 

We then write:
\begin{align}
\mathcal{J} &= \int_{\sbb}\exp\left(-\sum_{i=1}^m \sum_{j=1}^k\frac{({\lambda^i_j})^2m}{2(\sigma_j)^2} (y^i_j - s_j)^2\right)p(\sbb)d\sbb \\
&= \int_{\sbb} \prod_{i=1}^m \prod_{j=1}^k \exp\left(-\frac{({\lambda^i_j})^2m}{2(\sigma_j)^2}(y^i_j - s_j)^2 \right) \prod_{j=1}^k \delta(s_j) d\sbb \\
&= \prod_{j=1}^k \left[\int_{s_j} \exp\left(- \sum_i\frac{({\lambda^i_j})^2m}{2(\sigma_j)^2}(y^i_j - s_j)^2 \right)\delta(s_j) ds_j\right],
\end{align}
where $y^i_j=(\wb_j^i)^{\top}\xb^i$. We denote $\tilde{s}_j=\sum_{i=1}^m ({\lambda^i_j})^2 y^i_j$.  Fix $j$, and drop it to simplify notation. Then we need to solve the integral
\begin{align}
   &\int_s \exp\left(- \sum_i\frac{{\lambda^i}^2m}{2\sigma^2}(y^i - s)^2 \right)\delta(s) ds\\
   &=\int_s \exp \left(-\frac{m}{2\sigma^2}\sum_i {\lambda^i}^2 (y^i - \tilde{s} + \tilde{s} - s)^2 \right) \delta(s)ds \\ 
   &=\int_s \exp \left(-\frac{m}{2\sigma^2}\sum_i {\lambda^i}^2 ((y^i - \tilde{s})^2 + (\tilde{s} - s)^2) \right) \delta(s)ds \\ 
    &=\exp \left(-\frac{m}{2\sigma^2}\sum_i {\lambda^i}^2 ((y^i - \tilde{s})^2) \right)\int_z \exp \left(-\frac{m}{2\sigma^2} z^2 \right)
\delta(\tilde{s}-z) dz \label{21}
\end{align}
where in~\ref{21}, we make the change of variable $z=\tilde{s}-s$. The remaining
integral simply means that $\delta$ is smoothed by a Gaussian kernel. We then
define 
$f(s_j, \sigma_j) = -\log \left(\int_z \exp \left(-\frac{m}{2(\sigma_j)^2} z^2
  \right) \delta(s_j-z) dz\right)$ and obtain:
\begin{equation}
-\log(\mathcal{J}) = \sum_{i, j} \frac{m}{2(\sigma_j)^2} ({\lambda^i_j})^2
(y^i_j - \tilde{s}_j)^2 + \sum_j f(\tilde{s}_j, \sigma_j)
\end{equation}
\subsection{Updates and formulas for gradients and Hessians of loss~\eqref{eq:avica_likelihood_unconstrained}}
\label{app:gradients}
\paragraph{Updates w.r.t. $\lambda^i_j$}
Following~\cite{absil2012projection}, we use the retraction
\begin{equation}
  R(\xb) = \frac{\xb}{\|\xb\|\sqrt{1 - m\mu^2}}
\end{equation}
The Riemannian gradient $G^{\etab_j}$ and Hessian $H^{\etab_j}$ of $\loss$ are defined by:
$
  \loss(\frac{\etab_j + \epsilon}{\|\etab_j + \epsilon\|\sqrt{1 - m\mu^2}}) = \loss(\etab_j) + \langle \epsilon,
  G^{\etab_j}\rangle + \frac12 \langle \epsilon | H^{\etab_j} \epsilon \rangle
  + o(\|\epsilon\|^2)
$
where $\|\etab_j\| = \sqrt{1 - m\mu^2}$.
Due to the above definitions we get after standard derivations the following relationship between the
Riemannian gradient $G^{\etab_j}$ and the Euclidean gradient $\gb^{\etab_j}$:
\begin{equation}
  G^{\etab_j} = \gb^{\etab_j} - \frac{\etab_j}{\|\etab_j\|} \langle \frac{\etab_j}{\|\etab_j\|} | \gb^{\etab_j} \rangle 
\end{equation}
 
The relationship between the Riemannian Hessian $H^{\etab_j}$ and the Euclidean
Hessian $\mathcal{H}^{\etab_j}$ is given by:
\begin{align*}
  H^{\etab_j} &= -\langle G^{\etab_j} | \frac{\etab_j}{\|\etab_j\|^2} \rangle I_m - G^{\etab_j} \left(\frac{\etab_j}{\|\etab_j\|^2}\right)^{\top} + 3\langle G^{\etab_j} | \frac{\etab_j}{\|\etab_j\|^2} \rangle \etab_j \left(\frac{\etab_j}{\|\etab_j\|^2}\right)^{\top} \\ &+ \left[I_m - \etab_j \left(\frac{\etab_j}{\|\etab_j\|^2}\right)^{\top}\right] \mathcal{H}^{\etab_j} \left[I_m - \etab_j \left(\frac{\etab_j}{\|\etab_j\|^2}\right)^{\top}\right] \numberthis
\end{align*}

The gradient $G^{\etab_j}$ and $H^{\etab_j}$ are therefore available in
closed-form and updates are given by
\begin{equation}
\etab_j \leftarrow R(\etab_j - \rho (H^{\etab_j})^{-1} G^{\etab_j})
\end{equation}  
which are the same updates as in~\eqref{update_l}.

\paragraph{Formulas for gradients and Hessians}
\begin{itemize}
  \item Formulas for $W^i$:
the gradient $G^{W^i}$ is given by
\begin{equation}
G^{W^i} = \left[({\lambdab^i})^2 \odot \phi'(\tilde{\sbb}, \sigma_j) \right] (\yb^i)^{\top} + m \left[\sigmab^{-2} \odot (\mathbf{1} -
({\lambdab^i})^2) \odot ({\lambdab^i})^2 \odot (\yb^i - (\tilde{\sbb}^{-i}) \odot (\mathbf{1} - ({\lambdab^i})^2) \right](\yb^i)^T - I_k
\end{equation}
where $\odot$ is the component-wise product and $\lambdab^i, \sigmab, \mathbf{1} \in \mathbb{R}^k$ are
such that $\lambdab^i[j] = \lambda^i_j$, $\sigmab[j] = \sigma_j^2$ and
$\mathbf{1}[j] = 1$ and $\tilde{\sbb}^{-i}= \sum_{z \neq i} (\lambdab^z)^2 \odot \yb^z$. The Hessian is given in the main text.

\item Formulas for $\eta_j$:
  We give here the Euclidean gradient $\gb^{\etab_j}$ and Hessian $\mathcal{H}^{\etab_j}$.
  \begin{equation}
  \gb^{\etab_j} =  -\etab_j \odot (\lambdab_j)^{-2} + \frac{m}{(\sigma_j)^2} \etab_j  \odot  (\stil_j\mathbf{1} - \yb_j)^2 + 2 \etab_j \odot \yb_j \phi'(\stil_j, \sigma_j) 
  \end{equation}
  
  \begin{align*}
    &\mathcal{H}^{\etab_j} = 8 \frac{m}{2 \sigma_j^2} \left(\etab_j \odot (\stil_j \mathbf{1} - \yb_j)\right) \left(\yb_j \odot \etab_j \right)^{\top}
      + 8 \frac{m}{2 \sigma_j^2} \left(\yb_j \odot \etab_j \right) \left(\etab_j \odot (\stil_j \mathbf{1} - \yb_j)\right)^{\top} \\
    &+ 8 \left(\yb_j \odot \etab_j \right)\left(\yb_j \odot \etab_j \right)^{\top} (\frac{m}{2 \sigma_j^2} + \frac12 \phi''(\stil_j, \sigma_j)) \\
    & + \diag \left[ 2 \etab_j^2 \odot \lambdab_j^{-4} + 2 \yb_j \phi'(\stil_j, \sigma_j) - \frac1{\lambdab_j^2} + 2\frac{m}{2 \sigma_j^2} (\stil_j \mathbf{1} - \yb_j)^2\right] 
  \end{align*}

  \item Formulas for $\sigma_j$: the gradient $G^{\sigma_j}$ and Hessian
    $H^{\sigma_j}$ are given by

    \begin{equation}
      G^{\sigma_j} = \frac{(m-1)}{\sigma_j} - \frac{m}{\sigma_j^3} \left(\sum_i (\lambda^i_j)^2 (y^i_j - \stil_j)^2\right) + \partialfrac{\sigma_j}{\phi}(\stil_j, \sigma_j)
    \end{equation}
    
    \begin{equation}
      H^{\sigma_j} = -\frac{(m-1)}{\sigma_j^2} + 3\frac{m}{\sigma_j^4} \left(\sum_i (\lambda^i_j)^2 (y^i_j - \stil_j)^2\right) + \partialfrac{\sigma_j^2}{\phi}(\stil_j, \sigma_j)
    \end{equation}

\end{itemize}

\section{Proofs}
\label{sec:app_proofs}

\subsection{Proof of Proposition~\ref{prop:identifiability}}
\label{app:identifiability}
We fix a subject $i$. Since $\sbb$ has independent non-Gaussian components, it
is also the case for $\sbb + \nb^i$. Following~\cite{comon1994independent},
Theorem 11, there exists a scale-permutation matrix $P^i$ such that $A'^i =
A^iP^i$. As a consequence, we have $\sbb  + \nb^i = P^i(\sbb' + \nb'^i)$ for all
$i$.
Then, we focus on subject 1 and subject $i \neq 1$:
\begin{align}
  &\sbb + \nb^1 - (\sbb + \nb^i) = P^1(\sbb' + \nb'^1) - P^i(\sbb' + \nb'^i)\\
  &\nb^1 - \nb^i = P^1(\sbb' + \nb'^1) - P^i(\sbb' + \nb'^i)\\
  &\iff P^1\sbb' - P^i\sbb' = P^i \nb'^i - \nb^i + \nb^1 - P^1 \nb'^1
\end{align}
This shows that $P^1\sbb' - P^i\sbb'$ is either zero or Gaussian which can only happen if $P^1
= P^i$ (following Lemma 9 of~\cite{comon1994independent}).
Therefore, the matrices $P^i$ are all equal, and there exists a scale and permutation matrix $P$ such that $A'^i = A^iP$.

Now we also have

\begin{align}
  &\sbb + \nb^i  = P(\sbb' + \nb'^i) \\
  &\iff \sbb - P\sbb' = P\nb'^i - \nb^i
\end{align}

By a similar argument as before we get $\sbb = P\sbb'$ and
$\nb^i = P{\nb^i}'$ and therefore $\Sigma^i = P {\Sigma^i}' P^\top$.

  \subsection{Proof of Proposition~\ref{prop:robustness}}
  \label{app:robustness}
We denote ${W^i}^* = \Gamma^i (A^i)^{-1}$, ${(\yb^i)}^* = {W^i}^* \xb^i = \Gamma^i ({\sbb^i}^*
+ {\nb^i}^*)$ and introduce $\gamma^i_j$ such that $\Gamma^i = \diag(\gamma^i_1 \dots
\gamma^i_k)$. 

The model of AVICA becomes:
\begin{align*}
  & y^i_j = \frac1{\gamma^i_j}(s_j + n^i_j), \enspace i=1,\dots, m \\
  &p(\sbb_j) = \delta(s_j) \\
  &n^i_j \sim \mathcal{N}(0, {\sigma^i_j}^2)
  \numberthis \label{simplikelihood}
\end{align*}
where ${\sigma^i_j}^2 = \frac{(\sigma_j)^2}{m({\lambda^i_j})^2}$ and $\lambda^i_j$
respects the constraints introduced in ($\mathcal{H}$) and $\sum_{i=1}^m
(\lambda^i_j)^2 = 1$ for all $j$. 

Note that there are no interactions between components so the derivations can
be performed with only one component (which we omit in the notations for simplicity):
\begin{align}
  p({(y^1)}^* \cdots {(y^m)}^*) &= \int_s \prod_{i=1}^m p({(y^i)}^*|s)\delta(s) ds \\
&\leq \int_s p({(y^i)}^*|s)\delta(s) ds \\
                        &=  |\gamma^i| \int_s \mathcal{N}(s; \gamma^i{(y^i)}^*, \sigma^i) \delta(s) ds \label{smallratio2} \\
                        &=  |\gamma^i| \int_s \frac{\exp(-\frac{(\gamma^i{(y^i)}^* - s)^2}{2 {\sigma^i}^2})}{\sqrt{2 \pi{\sigma^i}^2}} \delta(s) ds \label{smallratio}
\end{align}

From equation~\ref{smallratio} we use $\exp(-\frac{(\gamma^i{(y^i)}^* - s)^2}{2
  {\sigma^i}^2}) \leq 1$ which gives $p({y^1}^* \cdots {y^m}^*) \leq
\frac{|\gamma^i|}{\sqrt{2 \pi}|\sigma^i|}$ and therefore denoting $\mathcal{L} = -\log(p({y^1}^* \cdots
  {y^m}^*))$ the negative log-likelihood, we have
\begin{equation}
\label{33}
\lim_{\frac{|\gamma^i|}{|\sigma^i|} \rightarrow 0} \mathcal{L} = +\infty
\end{equation}
for all $i$.

From equation~\ref{smallratio2} we use $\mathcal{N}(s; \gamma^i{(y^i)}^*,
\sigma^i) \leq 1$ which gives $p({(y^1)}^* \cdots { y^m }^*) \leq |\gamma^i|$
and therefore

\begin{equation}
\label{34}
\lim_{|\gamma^i| \rightarrow 0} \mathcal{L} = +\infty
\end{equation}
for all $i$.

We also have

\begin{align}
  &p({(y^1)}^* \cdots {(y^m)}^*) \\
  &= \int_s \prod_{i=1}^m p({(y^i)}^*|s)\delta(s) ds \\
&= \int_s \prod_{i=1}^m \left[|\gamma^i| \int_s \frac{\exp(-\frac{(\gamma^i{(y^i)}^* - s)^2}{2 {\sigma^i}^2})}{\sqrt{2 \pi{\sigma^i}^2}} \delta(s) \right]ds \\
&= \int_s \prod_{i=1}^m |\gamma^i| \int_s \frac{\exp(-\frac{\sum_i m(\lambda^i)^2(\gamma^i{(y^i)}^* - s)^2}{2 \sigma^2})}{\sqrt{2 \pi\frac{(\sigma)^2}{m(\lambda^i)^2}}} \delta(s) ds \label{37}\\
&\propto \int_s \prod_{i=1}^m \left[\sqrt{\frac{(\gamma^i)^2 m (\lambda^i)^2}{\sigma^2}} \right] \int_s \exp(-\frac{ m}{2 \sigma^2}\sum_i\left[(\lambda^i)^2(\gamma^i{(y^i)}^* - s)^2\right]) \delta(s) ds \label{38}\\
&= \int_s \prod_{i=1}^m \left[\sqrt{\frac{(\gamma^i)^2 m (\lambda^i)^2}{\sigma^2}} \right] \int_s \exp(-\frac{ m}{2 \sigma^2}\sum_i\left[(\lambda^i)^2\left((\gamma^i{(y^i)}^* - \stil^*)^2 + (\stil^* - s)^2\right)\right]) \delta(s) ds \label{39}\\
&= \int_s \prod_{i=1}^m \left[\sqrt{\frac{(\gamma^i)^2 m (\lambda^i)^2}{\sigma^2}} \right] \int_s \exp(-\frac{ m}{2 \sigma^2}\sum_i\left[(\lambda^i)^2(\gamma^i{(y^i)}^* - \stil^*)^2 \right]) \exp(-\frac{ m}{2 \sigma^2}(\stil^* - s)^2) \delta(s) ds \label{40}\\
&\leq \prod_{i=1}^m \left[\sqrt{\frac{(\gamma^i)^2 m (\lambda^i)^2}{\sigma^2}} \right] \exp(-\frac{ m}{2 \sigma^2}\sum_i\left[(\lambda^i)^2(\gamma^i{(y^i)}^* - \stil^*)^2 \right])\label{41}\\
&= \prod_{i=1}^m \left[\sqrt{\frac{(\gamma^i)^2 m (\lambda^i)^2}{\sigma^2}} \exp(-\frac{ m}{2 \sigma^2}(\lambda^i)^2(\gamma^i{(y^i)}^* - \stil^*)^2) \right]\label{42}
\end{align}
where in equation~\ref{38}, we use the parametrization ${\sigma^i}^2 =
\frac{(\sigma)^2}{m({\lambda^i})^2}$ where $\sum_i (\lambda^i)^2 = 1$ and use $\stil^*
= \sum_i ( \lambda^i )^2 \gamma^i {(y^i)}^*$. In equation~\ref{41}, we used
$\exp(-\frac{ m}{2 \sigma^2}(\stil^* - s)^2) \leq 1$.

The negative log-likelihood therefore verifies:
\begin{align}
  &\loss \geq \sum_{i=1}^m \left[-\frac12 \log(\frac{(\gamma^i)^2 m (\lambda^i)^2}{\sigma^2}) +  \frac{ m}{2 \sigma^2} (\lambda^i)^2(\gamma^i{(y^i)}^* - \stil^*)^2 \right]
\end{align}

Let us focus on the term $(\gamma^i{(y^i)}^* - \stil^*)^2$:
\begin{align}
  (\gamma^i{(y^i)}^* - \stil^*)^2 &= \left(\gamma^i ({ s }^* + { n^i }^*) - \sum_{z=1}^m \gamma^z ({ s }^* + { n^z }^*) {\lambda^z}^2\right)^2 \\
  &=\left(\gamma^i - \sum_{z=1}^m \gamma^z (\lambda^z)^2\right)^2 { s^* }^2  +\left(\gamma^i { n^i }^* - \sum_{z=1}^m \gamma^z (\lambda^z)^2{ n^z }^* \right)^2 \\
  &\geq \left(\gamma^i { n^i }^* - \sum_{z=1}^m \gamma^z (\lambda^z)^2{ n^z }^* \right)^2 \\
  &= \left(\gamma^i (1 - (\lambda^i)^2){ n^i }^* - \sum_{z=1, z \neq i}^m \gamma^z (\lambda^z)^2{ n^z }^* \right)^2 \\
  &\geq \left(\gamma^i (1 - (\lambda^i)^2){ n^i }^*\right)^2 \geq  (\gamma^i)^2c^i
\end{align}
where $c^i > 0$ is a strictly positive constant because of $( \mathcal{H} )$.

Therefore
\begin{align}
  &\loss \geq \sum_{i=1}^m \left[-\frac12 \log(\frac{(\gamma^i)^2 m (\lambda^i)^2}{\sigma^2}) +  \frac{ m}{2 \sigma^2} (\lambda^i)^2(\gamma^i)^2 c^i \right] \\
  &= \sum_{i=1}^m \left[-\frac12 \log(\frac{(\gamma^i)^2}{(\sigma^i)^2}) +  \frac{(\gamma^i)^2}{2 (\sigma^i)^2}  c^i \right]
\end{align}

We then have
\begin{equation}
\label{52}
\lim_{\frac{|\gamma^i|}{|\sigma^i|} \rightarrow \infty} \mathcal{L} = +\infty
\end{equation}
for all $i$.

Using~\eqref{52} and~\eqref{34} yields
\begin{equation}
\label{53}
\lim_{|\sigma^i| \rightarrow 0} \mathcal{L} = +\infty
\end{equation}
for all $i$.

We now study what happens when $\forall i, \gamma^i \rightarrow \infty$ while
$ \forall i, r^i = \frac{|\sigma^i|}{|\gamma^i|}$ is bounded.

The model becomes
\begin{align*}
  & y^i = a^is + n^i, \enspace i=1,\dots, m \\
  &p(s) = \delta(s) \\
  &n^i \sim \mathcal{N}(0, (r^i)^2)
\end{align*}
in the limit $a^i \rightarrow 0$ for all $i$.

The log likelihood can be written:

\begin{align}
  h((a^i)_{i=1}^m) &= \log p(y^1 \dots y^m) \\
  &= \log \int_s \prod_i \mathcal{N}(y^i - a^is; 0, (r^i)^2) p(s) ds
\end{align}

We then have
\begin{align}
  &\partialfrac{a^i}{h} = (\frac1{h} \partialfrac{a^i}{h}) \\
                       &= \frac{\int_s -s\frac{(y^i - a^i s)}{{r^i}^2} \prod_z \mathcal{N}(y^z - a^zs; 0, (r^z)^2) p(s)ds }{\int_s \prod_i \mathcal{N}(y^i - a^is; 0, (r^i)^2) p(s)ds}
\end{align}
Therefore if $\forall i, a^i = 0$, we get:

\begin{align}
  &\partialfrac{a^i}{h}(0) 
  = \frac{\int_s -s\frac{(y^i)}{{r^i}^2} p(s) ds }{\int_s p(s) ds} = 0
\end{align}
since $\int_s sp(s) ds = 0$.

Therefore a stationary point is reached when the scale goes to infinity. We show
that this stationary point is not a maximum of the log-likelihood.

\begin{align}
  \partialfrac{a^i a^j}{h}(0) &= (\frac1{h} \partialfrac{a^i a^j}{h} - \frac1{h^2} \partialfrac{h^i}{p} \partialfrac{h^j}{p})(0) \\
                              &= (\frac1{h} \partialfrac{a^i a^j}{h})(0) 
\end{align}

If $i \neq j$ we get:

\begin{align}
  &\frac1{h} \partialfrac{a^i a^j}{h} = \\
  &\frac{\int_s s^2\frac{(y^i - a^i s)}{{r^i}^2} \frac{(y^j - a^j s)}{{r^j}^2}\prod_z \mathcal{N}(y^z - a^zs; 0, (r^z)^2 ) p(s)ds }{\int_s \prod_z \mathcal{N}(y^z - a^z s ; 0, (r^z)^2) p(s)ds} \\
  & \implies  (\frac1{h} \partialfrac{a^i a^j}{h})(0) = \frac{y^i y^j}{(r^i r^j)^2}
\end{align}
since $\int_s s^2 p(s) ds = 1$.

If $i = j$ we get:
\begin{align}
  &\frac1{h} \partialfrac{a^i a^i}{h} \\
  &= \frac{\int_s (s^2\frac{(y^i - a^i s)^2}{(r^i)^4} + \frac{2a^is - y^i}{{r^i}^2})\prod_z \mathcal{N}(y^z - a^z s ; 0, (r^z)^2) p(s)ds }{\int_s \prod_z \mathcal{N}(y^z - a^zs ; 0, (r^z)^2) p(s)ds} \\
  & \implies  (\frac1{h} \partialfrac{a^i a^i}{h})(0) = \frac{(y^i)^2}{(r^i)^4} - \frac{y^i}{{r^i}^2}
\end{align}
since $\int_s s^2 p(s) ds = 1$.

With $a^i=0$, $y^i$ has variance $(r^i)^2$ and zero mean. So, in
expectation the Hessian is positive definite.
Therefore $\forall i, a^i = 0$ is a minimum of the loglikelihood and a maximum of the
negative log-likelihood.

Therefore at the minima, there exists a finite $\gamma^i$. Therefore
$\sigma^i$ is also finite since the ratio must remain bounded. But since the
precisions are constrained to be in $[\mu^2, 1 - m \mu^2]$ if $\sigma^i$ is
finite $\forall i, \sigma^i$ is finite. Therefore $\forall i, \gamma^i$ is
finite since the ratio must remain bounded.

This shows that as the parameters get close to the border of the definition set of
loss~\eqref{eq:avica_likelihood_unconstrained}, the loss either goes to $+\infty$ or goes
towards a local maximum. Therefore, there exists parameters
$(\gamma^i_j)_{i=1, j=1}^{m, k}, (\lambda^i_j)_{i=1, j=1}^{m, k},
(\sigma_j)_{j=1}^k$ such that $(\Gamma^i (A^i)^{-1})_{i=1}^m, (\lambda^i_j)_{i=1, j=1}^{m, k},
(\sigma_j)_{j=1}^k$ is a well defined local minima of the loss~\eqref{eq:avica_likelihood_unconstrained}.

\section{Closed form MMSE estimator: detailed computations}
\label{appendix:mmse_derivation}

\begin{align*}
  &p(\xb, \sbb) = p(\xb | \sbb) p(\sbb) = \prod_{i=1}^m p(\xb^i | \sbb) p(\sbb) \\ &\propto \exp\left(-\sum_{i=1}^m \sum_{j=1}^k\frac{({\lambda^i_j})^2m}{2(\sigma_j)^2} (y^i_j - s_j)^2\right)p(\sbb) \\
  &\propto \prod_{i=1}^m \prod_{j=1}^k \left[\exp \left(-\frac{({\lambda^i_j})^2m}{2(\sigma_j)^2}(y^i_j - s_j)^2 \right)\delta(s_j) \right] \\
       &=\prod_{j=1}^k \exp \left(-\frac{m}{2(\sigma_j)^2}\sum_i ({\lambda^i_j})^2 ((y^i - \tilde{s}_j)^2) \right) \exp \left(-\frac{m}{2(\sigma_j)^2} (\tilde{s}_j - s_j)^2 \right) \delta(s_j) \\
                        &\propto \prod_{j=1}^k \exp \left(-\frac{m}{2(\sigma_j)^2} (\tilde{s}_j - s_j)^2 \right) \delta(s_j)
\end{align*}

We then have
\begin{align}
  &\exp \left(-\frac{m}{2(\sigma_j)^2} (\tilde{s}_j - s_j)^2 \right) \delta(s_j) \\
  &= \sqrt{2 \pi \frac{\sigma^2}{m}} \mathcal{N}(s_j; \tilde{s_j}, \frac{(\sigma_j)^2}{m}) \sum_{\alpha \in \{ \frac12, \frac32 \}} \mathcal{N}(s_j, 0, \alpha) \\
  & = \sqrt{2 \pi \frac{\sigma^2}{m}}  \sum_{\alpha \in \{ \frac12, \frac32 \}} \mathcal{N}(s_j; \tilde{s_j}, \frac{(\sigma_j)^2}{m}) \mathcal{N}(s_j, 0, \alpha) \\
  &=  \sqrt{2 \pi \frac{\sigma^2}{m}} \sum_{\alpha \in \{\frac12, \frac32\}}\mathcal{N}(\tilde{s_j}; 0, \alpha + \frac{\sigma^2}{m}) \mathcal{N}(s_j; \frac{\alpha \tilde{s}_j}{\alpha + \frac{(\sigma_j)^2}{m}}, \frac{\frac{(\sigma_j)^2}{m} \alpha}{\alpha + \frac{(\sigma_j)^2}{m}}) \label{78} \\
  &\propto \sum_{\alpha \in \{\frac12, \frac32\}}\mathcal{N}(\tilde{s_j}; 0, \alpha + \frac{\sigma^2}{m}) \mathcal{N}(s_j; \frac{\alpha \tilde{s}_j}{\alpha + \frac{(\sigma_j)^2}{m}}, \frac{\frac{(\sigma_j)^2}{m} \alpha}{\alpha + \frac{(\sigma_j)^2}{m}})
\end{align}

where we use in~\eqref{78} the fact that:
\begin{equation}
  \mathcal{N}(x; y, \nu) \mathcal{N}(x, 0, \alpha) = \mathcal{N}(y; 0, \nu + \alpha) \mathcal{N}(x; \frac{\alpha y}{\alpha + \nu}, \frac{\nu \alpha}{\alpha + \nu})
  \label{82}
\end{equation}

We now prove~\ref{82}:
\begin{align}
  \mathcal{N}(x; y, \nu) \mathcal{N}(x, 0, \alpha) &= \frac{\exp \left(-\frac{(x-y)^2}{2\nu} \right)}{\sqrt{2 \pi \nu}} \frac{\exp \left(-\frac{x^2}{2\alpha}\right)}{\sqrt{2 \pi \alpha}} \\
  &= \frac1{\sqrt{2 \pi \nu}\sqrt{2 \pi \alpha}}\exp\left(- \frac{\alpha (x-y)^2 + \nu x^2}{2 \alpha \nu} \right) \\
  &= \frac1{\sqrt{2 \pi \nu}\sqrt{2 \pi \alpha}}\exp\left(- \frac{\alpha (x^2 -2xy + y^2) + \nu x^2}{2 \alpha \nu} \right) \\
  &= \frac1{\sqrt{2 \pi \nu}\sqrt{2 \pi \alpha}}\exp\left(- \frac{x^2(\alpha + \nu) -2x(\alpha y) + \alpha y^2 }{2 \alpha \nu} \right) \\
  &= \frac1{\sqrt{2 \pi (\nu + \alpha)}\sqrt{2 \pi \frac{\nu \alpha}{\nu + \alpha}}}\exp\left( -\frac{x^2 -2x\frac{\alpha y}{\alpha + \nu} + \frac{\alpha y^2}{\alpha + \nu} }{2 \frac{\alpha \nu}{\alpha + \nu}} \right) \\
  &= \frac1{\sqrt{2 \pi (\nu + \alpha)}\sqrt{2 \pi \frac{\nu \alpha}{\nu + \alpha}}}\exp\left( -\frac{(x - \frac{\alpha y}{\alpha + \nu})^2 - ( \frac{\alpha y}{\alpha + \nu} )^2 + \frac{\alpha y^2}{\alpha + \nu} }{2 \frac{\alpha \nu}{\alpha + \nu}} \right) \\
  &= \frac1{\sqrt{2 \pi (\nu + \alpha)}\sqrt{2 \pi \frac{\nu \alpha}{\nu + \alpha}}}\exp\left( -\frac{(x - \frac{\alpha y}{\alpha + \nu})^2}{2\frac{\alpha \nu}{\alpha + \nu}}\right) \exp \left(-\frac{ - \alpha^2 y^2 + (\alpha + \nu)\alpha y^2 }{2 \alpha \nu(\alpha + \nu)} \right) \\
  &= \frac1{\sqrt{2 \pi (\nu + \alpha)}\sqrt{2 \pi \frac{\nu \alpha}{\nu + \alpha}}}\exp\left( -\frac{(x - \frac{\alpha y}{\alpha + \nu})^2}{2\frac{\alpha \nu}{\alpha + \nu}}\right) \exp \left(-\frac{\nu\alpha y^2 }{2 \alpha \nu(\alpha + \nu)} \right) \\
  &= \mathcal{N}(x; \frac{\alpha y}{\alpha + \nu}; \frac{\alpha \nu}{\alpha + \nu}) \mathcal{N}(y: 0, \alpha + \nu)
\end{align}

\section{Derivation of relative gradient and Hessian w.r.t $W^i$ in EM}
We use the relative gradient $\mathcal{G}^{W^i}$ and $\mathcal{H}^{W^i}$ defined
by $\mathcal{C}(W^i + \varepsilon W^i) = \mathcal{C}(W^i) + \langle
 \varepsilon|\mathcal{G}^{W^i}\rangle + \frac12 \langle
 \varepsilon|\mathcal{H}^{W^i}\rangle$.

 We get:
 \begin{align}
   \mathcal{C}(W^i + \varepsilon W^i) &= \sum_{i=1}^m \left[ -\log(|W^i|) -\log(|I_k + \varepsilon|) - \log(\mathcal{N}(\yb^i + \varepsilon \yb^i; \sbb; \Sigma^i)) \right] + const \\
                                      &= \mathcal{C}(W^i) - \tr(\varepsilon) + \frac12 \tr(\varepsilon^2) \\& \enspace \enspace + \frac12 \left[\langle \varepsilon \yb^i| (\Sigma^i)^{-1} (\yb^i - \sbb) \rangle + \langle (\yb^i - \sbb)| (\Sigma^i)^{-1} \varepsilon \yb^i \rangle + \langle \varepsilon \yb^i| (\Sigma^i)^{-1} \varepsilon \yb^i \rangle \right] + o(\|\varepsilon\|^2) \\
   &= \mathcal{C}(W^i) - \sum_a \varepsilon_{a, a} + \frac12 \sum_{a, b} \varepsilon_{a,b} \varepsilon_{b, a} \\& \enspace \enspace + \sum_{a, b} \varepsilon_{a, b} \left[(\Sigma^i)^{-1}(\yb^i - \sbb) (\yb^i)^{\top} \right]_{a, b} + \frac12 \sum_{a, b} \varepsilon_{a, b} \left[(\Sigma^i)^{-1} \varepsilon \yb^i (\yb^i)^{\top}\right]_{a, b} + o(\|\varepsilon\|^2) \\
   &= \mathcal{C}(W^i) - \sum_a \varepsilon_{a, a} + \frac12 \sum_{a, b} \varepsilon_{a,b} \varepsilon_{b, a} \\& \enspace \enspace + \sum_{a, b} \varepsilon_{a, b} \left[(\Sigma^i)^{-1}(\yb^i - \sbb) (\yb^i)^{\top} \right]_{a, b} + \frac12 \sum_{a, b, d} \varepsilon_{a, b} (\Sigma^i)^{-1}_{a, a} \varepsilon_{a, d} \left[\yb^i (\yb^i)^{\top}\right]_{d, b} + o(\|\varepsilon\|^2)
 \end{align}

 So:
 \begin{equation}
 \mathcal{G}^{W^i}_{a, b} =  -\delta_{a,b} + \left[(\Sigma^i)^{-1} (\yb^i - \sbb)(\yb^i)^{\top}\right]_{a, b}
 \end{equation}
 and
 \begin{equation}
 \mathcal{H}^{W^i}_{a, b, c, d} = \delta_{a, d}\delta_{b, c} + \delta_{a, c} \frac{y^i_b y^i_d}{\Sigma^i_a}
 \end{equation}

\label{gradients_em}

\section{Additional experiments}

\subsection{Further convergence plots}
\label{app:other_convergence_plot}
At each iteration, we record the estimated sources $\hat{\sbb}$.
The true sources $\sbb$ and the estimated sources are then normalized so that they
have unit variance. Then, using
the Hungarian algorithm they are matched and the sign is flipped so that for each
source $j$, $s_j$ and $\hat{s_j}$ have
maximum correlation.
Then, the reconstruction error is defined as $\frac1{k}\sum_{j=1}^k \left(1 -
  \mathbb{E}[s_j \hat{s_j}] \right)$.

\begin{figure}
  \centering
\includegraphics[scale=0.8]{"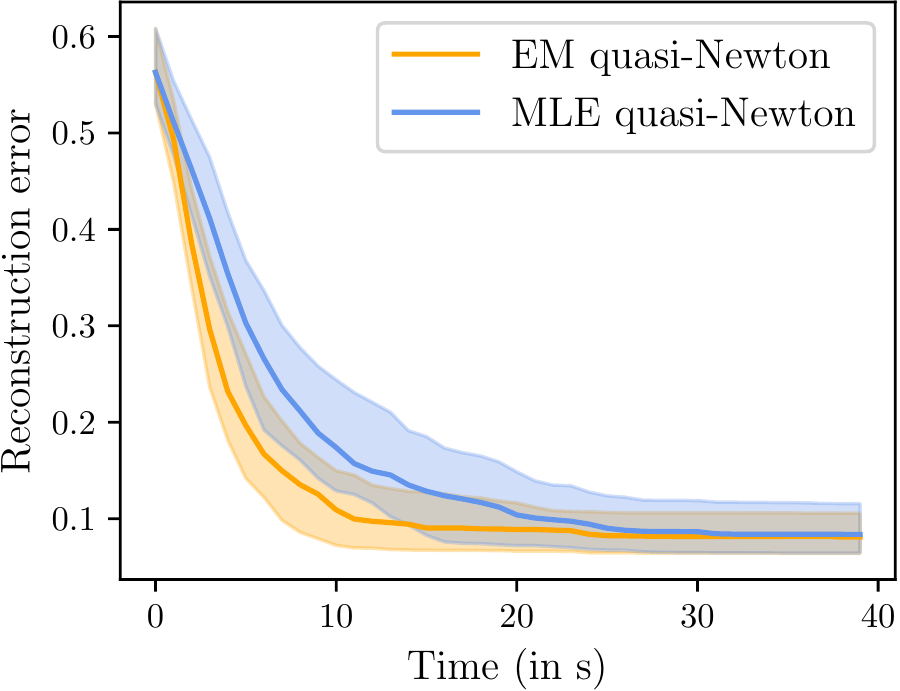"}
\label{add:conv_plot}
\caption{\textbf{Convergence plot}: Median reconstruction error as a function of
  time. Error bars display the first and last quartiles.}
\end{figure}

We run this experiment 100 times with different seeds.
We provide in Figure~\ref{add:conv_plot} the median reconstruction
error in function of time.

\subsection{Comparative study of the adaptive scaling of MVICA and AVICA}
\label{adaptivescaling}
Both MVICA and AVICA have the ability to put less weights on subjects that are
noisier. MVICA can do so by adjusting the scaling of the unmixing matrix of each subject while AVICA
can in addition make use of its estimate of the noise precisions.
Proposition~\ref{prop:identifiability} shows that AVICA is identifiable up to a
scaling and permutation. In contrast MVICA is identifiable up to a permutation
only as the noise variance is fixed to 1.
Therefore with MVICA there is a trade-off between scaling the data so that the
noise has variance 1 or scaling the data so that the sources have the correct
variance. Such issues do not occur in AVICA.

\begin{figure}
  \centering
  \includegraphics[width=0.5\textwidth]{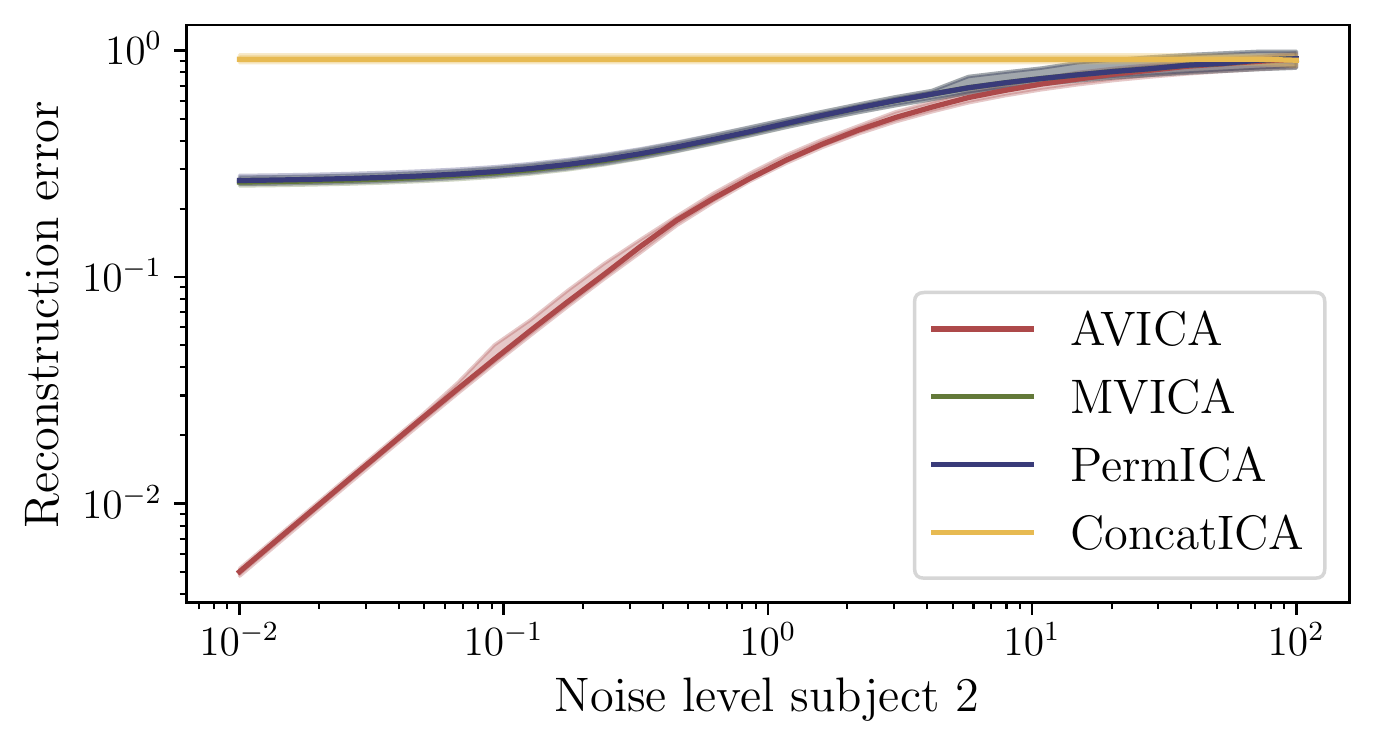}
  \caption{\textbf{Synthetic experiment - Adaptive scaling}: Reconstruction
    error as a function of the noise variance of subject 2 using only one component.
    The noise variance of subject 1 is fixed to 100. Error bars represent first and last deciles.}
  \label{fig:adaptive_scaling}
\end{figure}

In order to show this, we consider one-dimensional data $x^1$ and $x^2$ which
are both a scaled version of a common source. We add to subject $x^1$ a noise of
variance $100$ while we add a noise of variance between $10^{-2} \cdots 10^2$ to
the second subject. We apply ICA algorithms on these data and measure the reconstruction error as $1 - \mathbb{E}[s \hat{s}]$ between
the true source $s$ and the source $\hat{s}$ given by the algorithms after normalization to unit variance.
Results given in Fig~\ref{fig:adaptive_scaling} show that AVICA puts more
weight on the less noisy subject yielding a good estimate of the common sources
while other methods are not able to make use of the less noisy subjects.

\subsection{Parameter identification}
\label{paramident}
Adaptive multiViewICA MMSE estimator computes the shared response estimate as a weighted average of
individual responses where the weights corresponds to the
precisions. 

In this experiment we measure the ability of Adaptive multiViewICA to recover
the correct precisions using the same data as in the synthetic experiment in
section~\ref{synth_exp}.

\begin{figure}
  \centering
  \includegraphics[width=0.5\textwidth]{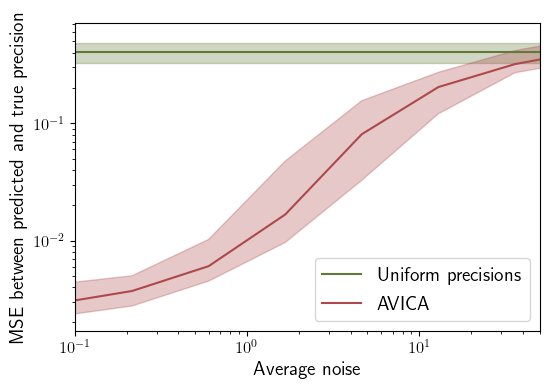}
  \caption{\textbf{Synthetic experiment - Error on precisions}: L2 distance between true and estimated
    precisions.}
  \label{fig:synth_param}
\end{figure}

We compute the reconstruction error between the true precisions ${\lambda^1_j}^2 \dots {\lambda^m_j}^2$ and the estimated precisions $\widehat{{\lambda^1_j}^2} \dots \widehat{{\lambda^m_j}^2}$
using the formula $\sum_{i=1}^m \sum_{j=1}^k (({\lambda^i_j})^2 - \widehat{({\lambda^i_j})^2})^2$ and show
the result in Figure~\ref{fig:synth_param} after applying the hungarian
algorithm to deal with permutation indeterminacies.

Figure~\ref{fig:synth_param} shows that the precisions obtained using Adaptive
multiViewICA are accurate. For comparison we give the reconstruction error
between the true precisions and the uniform assignment $(\lambda^i_j)^2 = \frac1{m}$.

\subsection{ROI chosen for the reconstruction experiment}
\label{app:roi}
\begin{figure}
  \centering
  \includegraphics[width=0.6\textwidth]{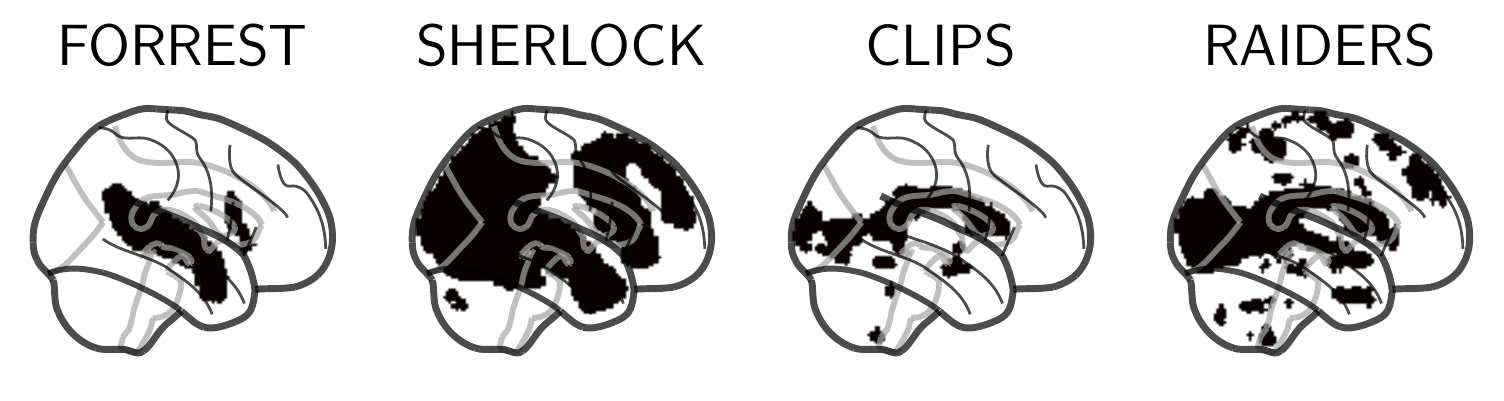}
  \caption{\textbf{ROI chosen in the reconstruction experiment in section~\ref{sec:fmri_exp}}}
  \label{fig:roi}
\end{figure}

We report in figure~\ref{fig:roi} the ROIs used in the reconstruction experiment in section~\ref{sec:fmri_exp}.

\subsection{Stability of spatial ICA decompositions on HCP Rest}
\label{fmristability}
In this experiment, we study the stability of spatial ICA decompositions on rest
fMRI data with respect to the group of subjects chosen to perform the
decomposition.
We use the HCP rest dataset~\cite{van2013wu}.

\begin{figure*}
  \centering
  \includegraphics[width=0.5\textwidth]{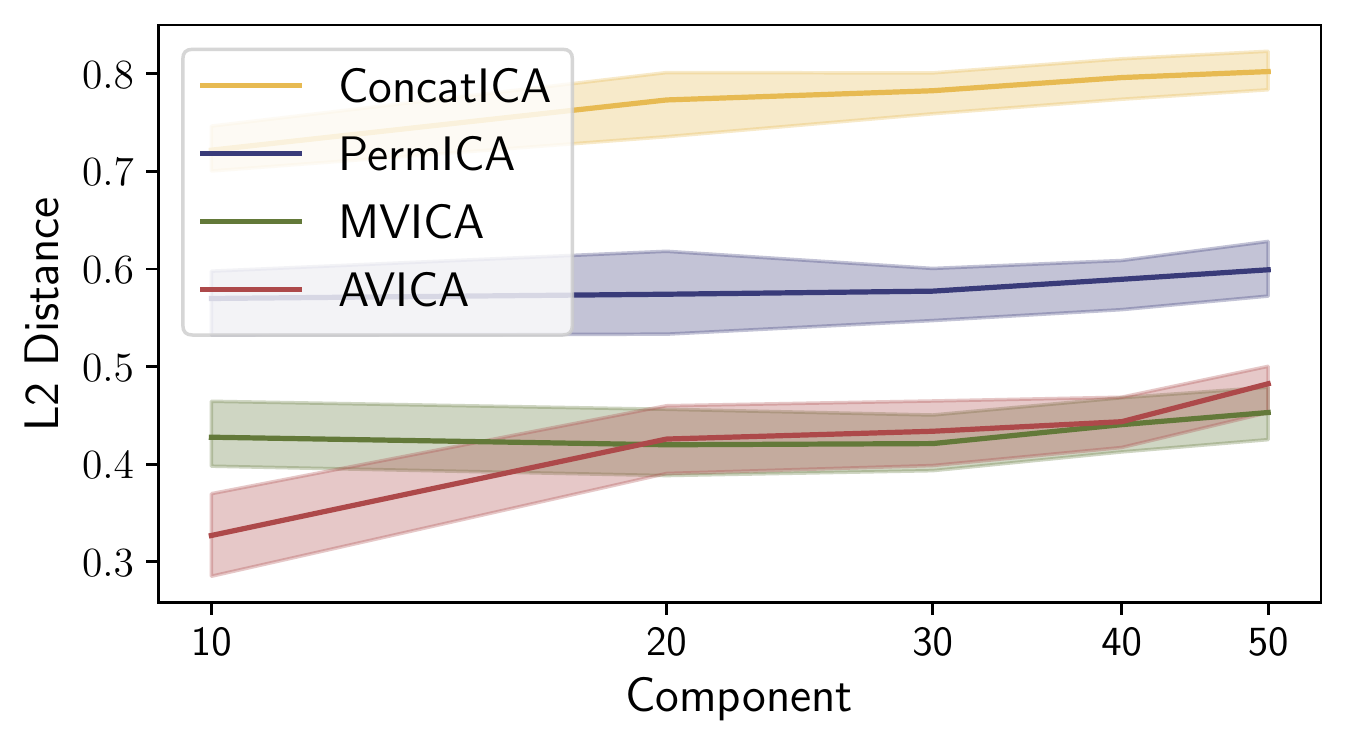}
  \caption{\textbf{Stability of spatial ICA decompositions on fMRI rest data}: We divide subjects into two groups of 100 sujects and perform spatial ICA
    separately. We match the components of the two groups and report the mean
    $\ell_2$ distance across components. The experiment is repeated 10 times. We report the median value. Error bars represent the lower and upper quartile.}
  \label{fig:stability}
\end{figure*}

We randomly divide subjects into two groups of 100 sujects and
perform spatial ICA separately. We match the components of the two groups using
the hungarian algorithm and compute the mean $\ell_2$ distance across components.
The experiment is repeated 10 times. We report
the median value in Figure~\ref{fig:stability}. Error bars represent the first and last quartiles.

We see that AVICA and MVICA yield more stable decompositions than other methods.
In particular, AVICA exhibits best performance with $10$ components. 

\subsection{MEG Phantom}
\label{megphantom}
We demonstrate the usefulness of AVICA on MEG data.
The first experiment uses data collected with a realistic head phantom, which is a plastic device mimicking real electrical brain sources.
Thirty two current dipoles positioned at different locations can be switched on or off.

We select $N_d$ dipole at random among the set of 32 dipoles where $N_d$ varies from 2 to 20. We view each dipole as a subject and therefore have $m=N_d$.
We only consider the 102 magnetometers.

An epoch corresponds to 3\,s of MEG signals where a dipole is switched on for 0.4\,s with an oscillation at 20\,Hz and a peak-to-peak amplitude of 200\,nAm.
This yields a matrix of size $p\times n$ where $p=102$ is the number of sensors, and $n$ is the number of time samples.
We have access to $100$ epochs per dipole.
For each dipole, we chose 2 epochs at random among the set of 100 epochs and concatenate them in the temporal dimension.

We apply algorithms to our data to extract $k=5$ shared sources.
As we know the true source (the timecourse of the dipole), we can compute the reconstruction error of each source as the squared norm of the difference between the estimated source and the true source, after normalization to unit variance and fixing the sign.
We only retain the source of minimal error.

\begin{figure}
  \centering
  \includegraphics[width=0.5\textwidth]{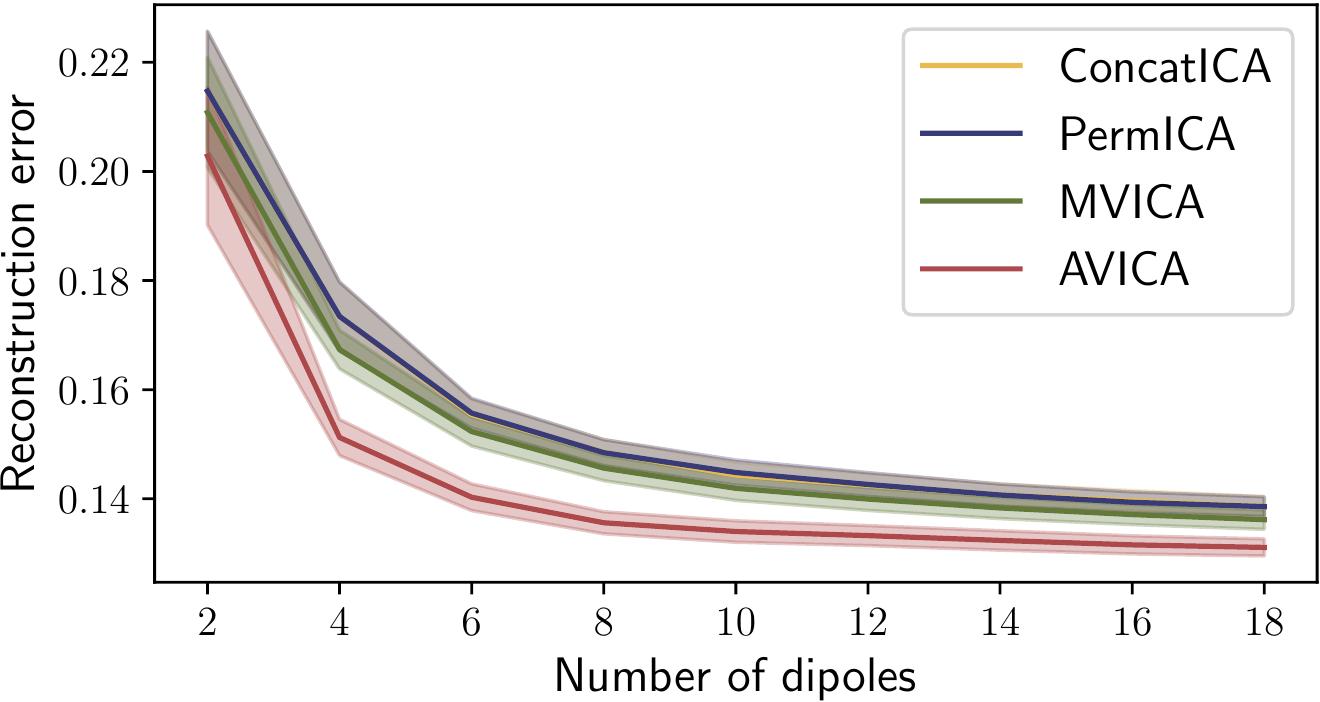}
  \caption{\textbf{MEG Phantom}: Reconstruction error between estimated and
    true source}
  \label{fig:eeg_phantom}
\end{figure}

This metric is reported in Figure~\ref{fig:eeg_phantom} when the number of dipoles considered $N_d$ varies.
Adaptive multiViewICA outperforms other methods.

\subsection{MEG CamCAN Timecourses}
\label{megsources}
\begin{figure}[H]
  \includegraphics[width=0.5\textwidth]{"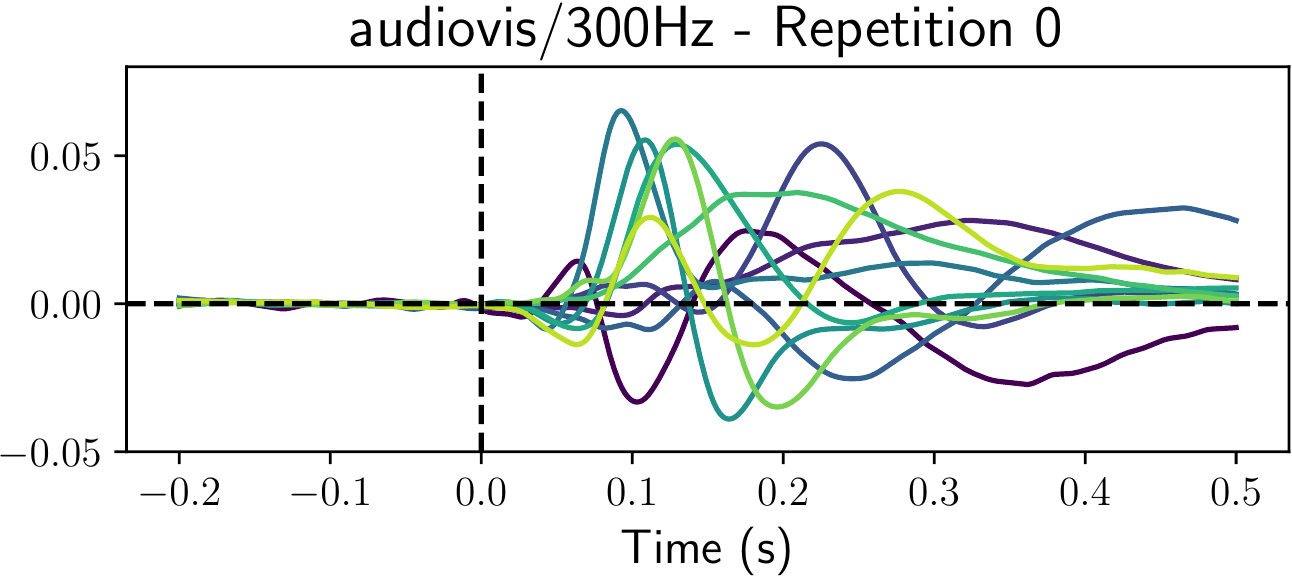"}
  \includegraphics[width=0.5\textwidth]{"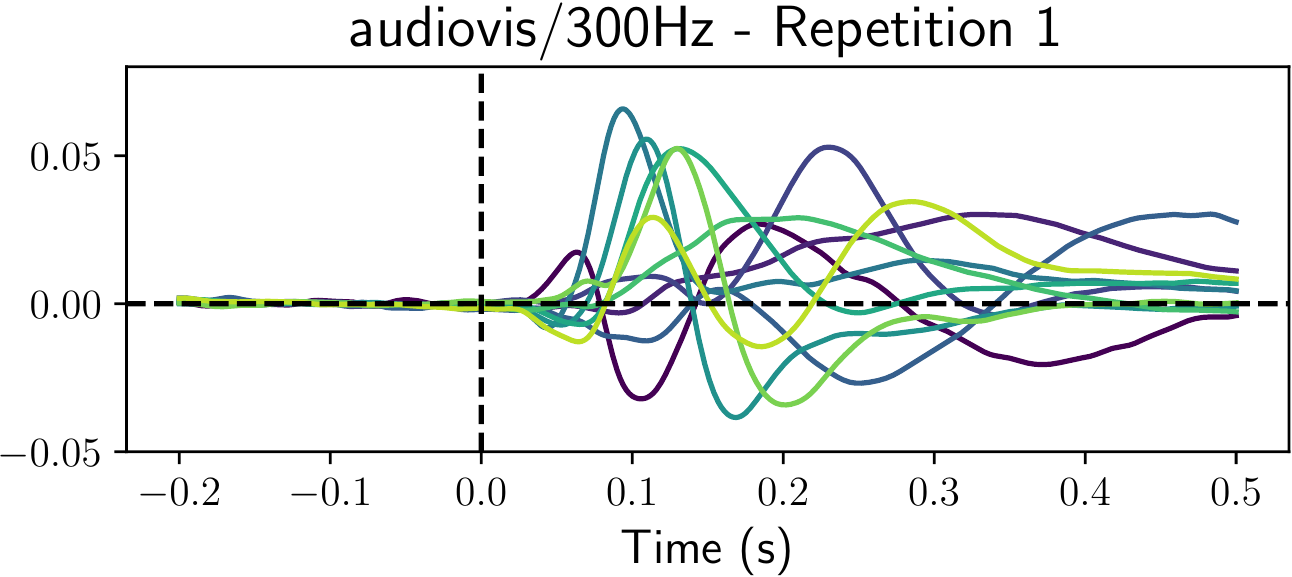"}
  \includegraphics[width=0.5\textwidth]{"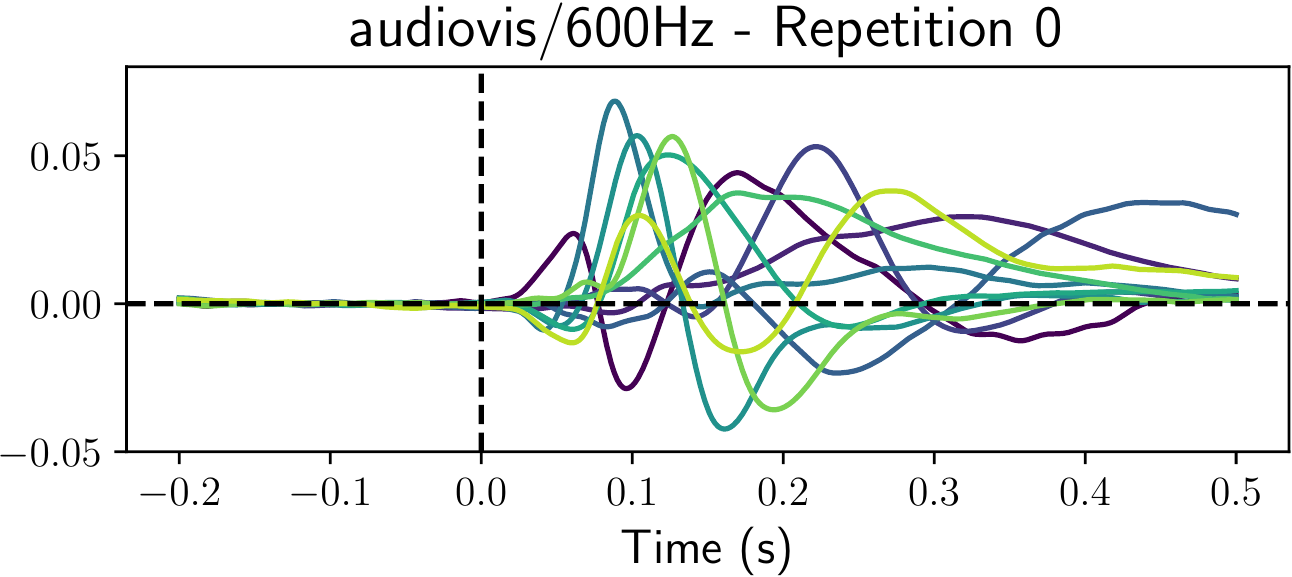"}
  \includegraphics[width=0.5\textwidth]{"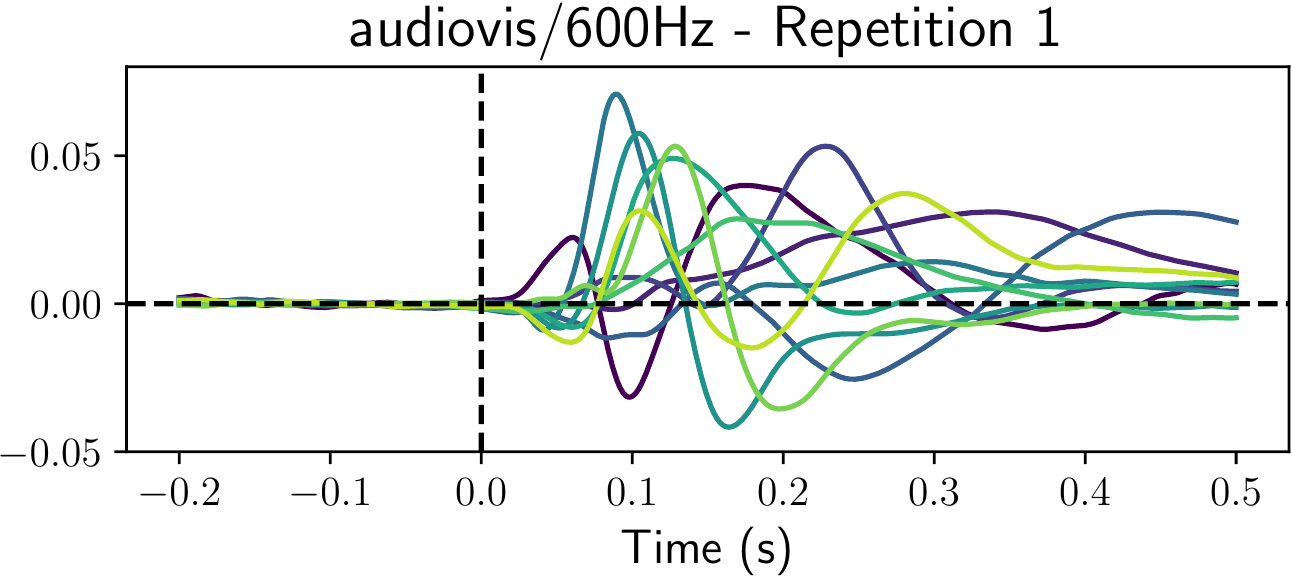"}
  \includegraphics[width=0.5\textwidth]{"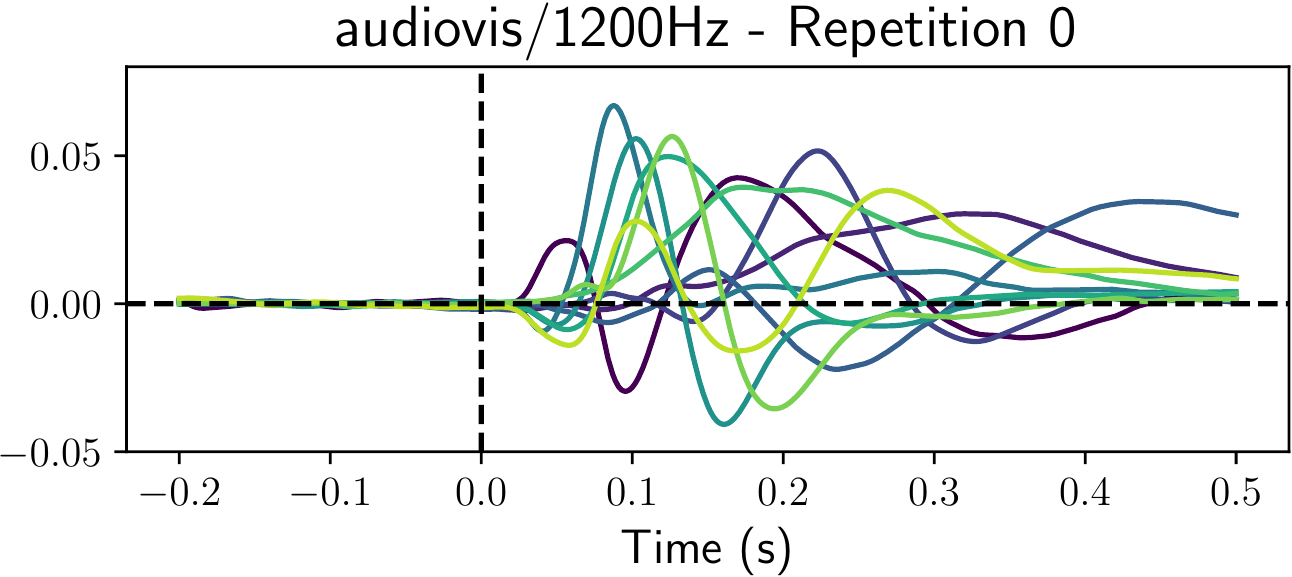"}
  \includegraphics[width=0.5\textwidth]{"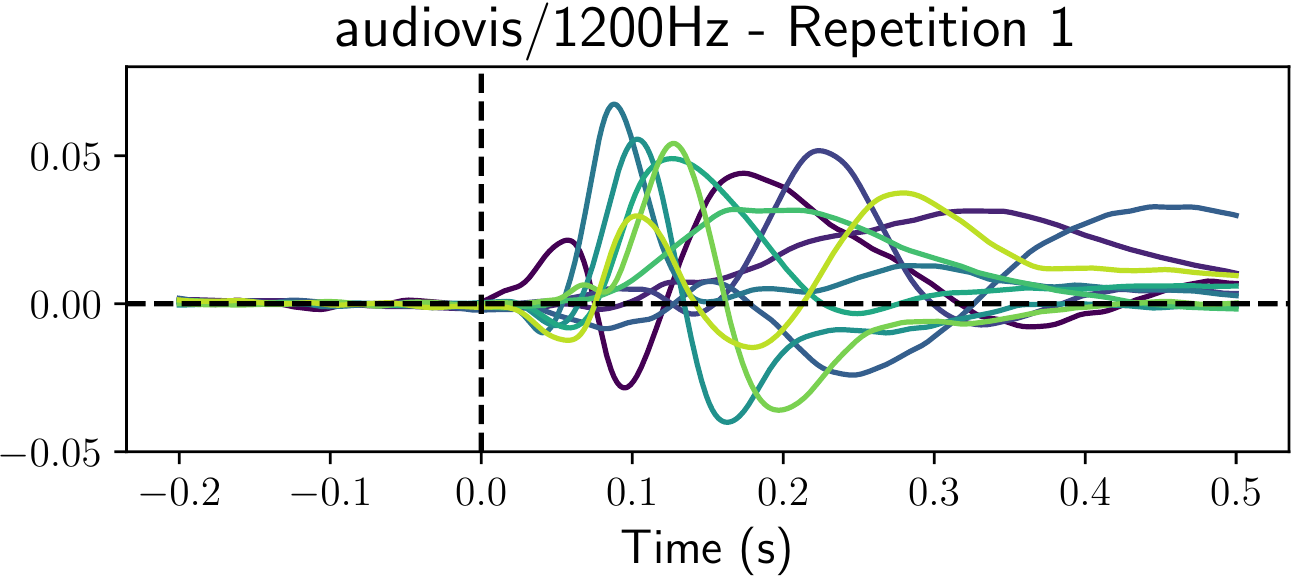"}
  \caption{\textbf{CamCAN MEG}: Timecourse of 10 shared sources recovered by
   Adaptive multiViewICA}
  \label{fig:eeg_source}
\end{figure}

We display in Figure~\ref{fig:eeg_source} the sources corresponding to the 6
chunks of data.

As we can see, the sources corresponding to the same stimuli are very similar over repetitions.

\subsection{Detailed CamCAN sources}
\label{megsourceslocal}

For each subject, sources are localized using the sLORETA algorithm~\cite{pascual2002standardized}
Then, they are registered to a common reference brain and averaged across subjects.
We plot the sources obtained using Adaptive multiViewICA  and their
localization below.

{\centering
  \includegraphics[width=0.3\textwidth]{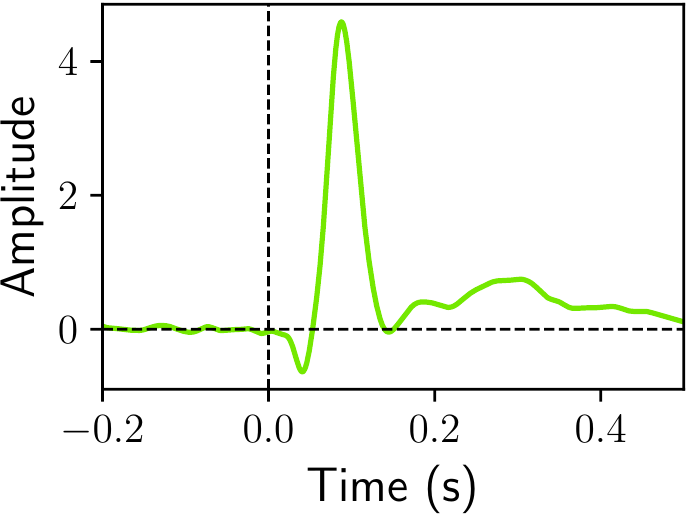}%
\raisebox{0.2\height}{\includegraphics[width=0.68\textwidth]{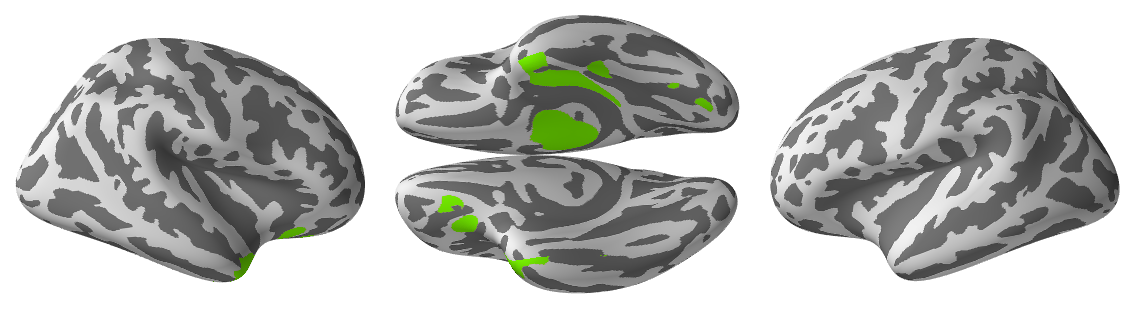}} \\
\includegraphics[width=0.3\textwidth]{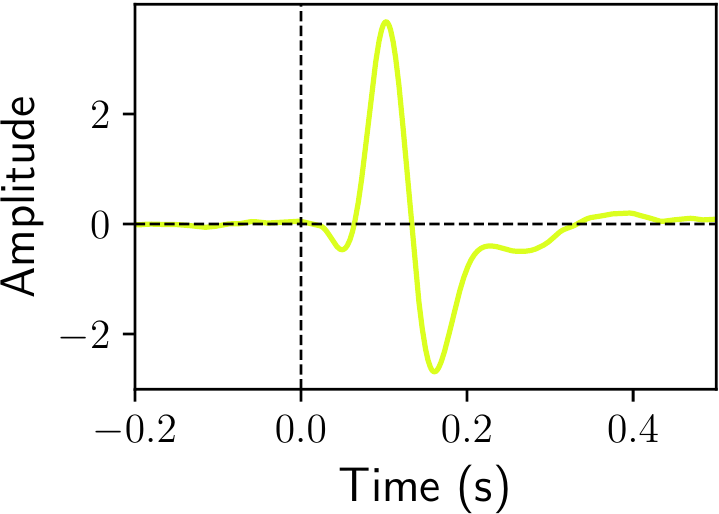}%
\raisebox{0.2\height}{\includegraphics[width=0.68\textwidth]{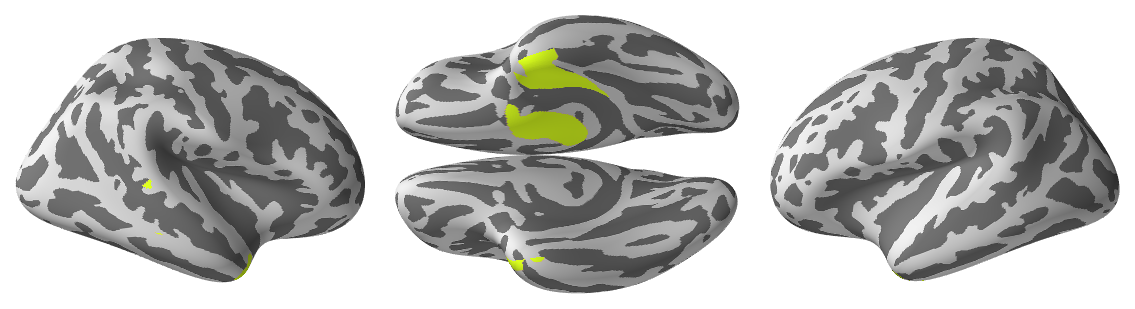}} \\
\includegraphics[width=0.3\textwidth]{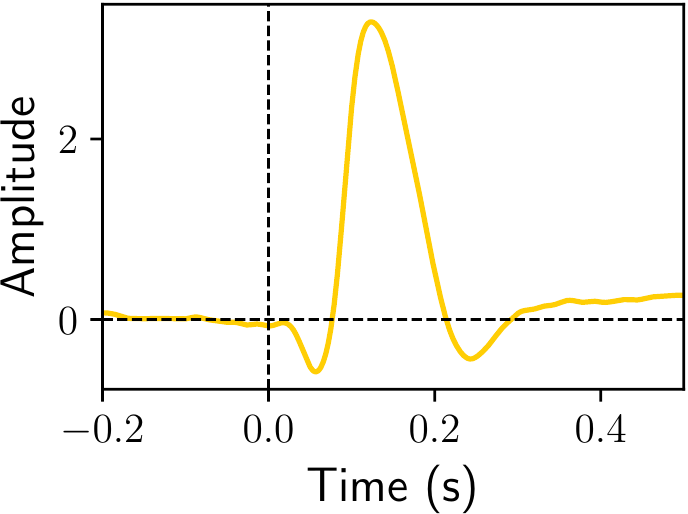}%
\raisebox{0.2\height}{\includegraphics[width=0.68\textwidth]{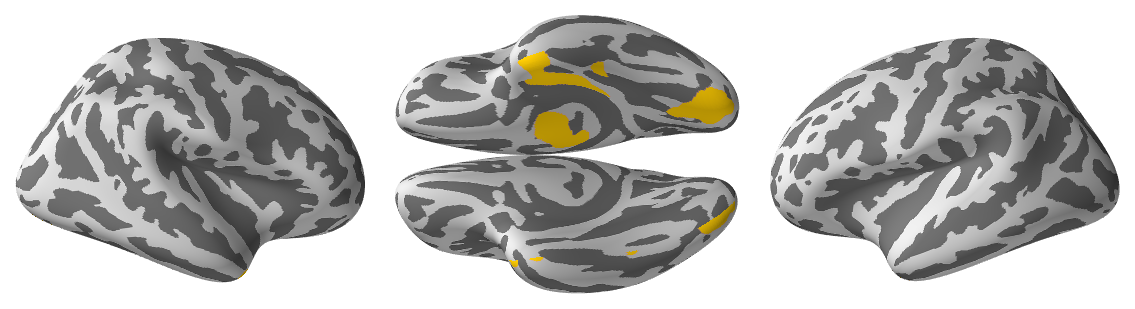}} \\
\includegraphics[width=0.3\textwidth]{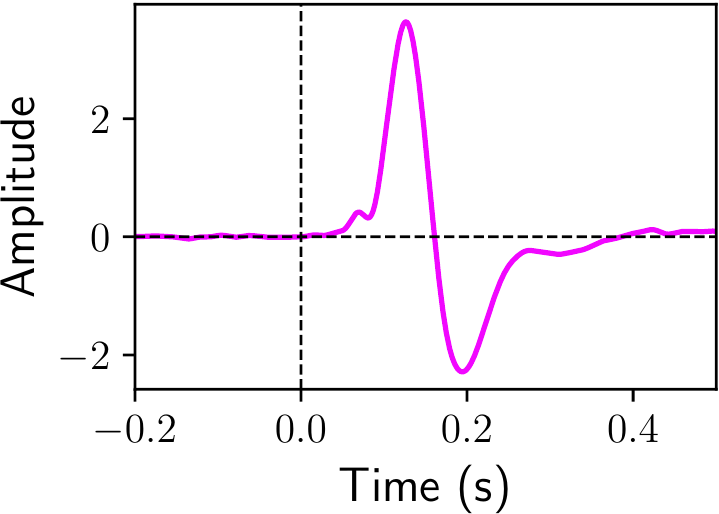}%
\raisebox{0.2\height}{\includegraphics[width=0.68\textwidth]{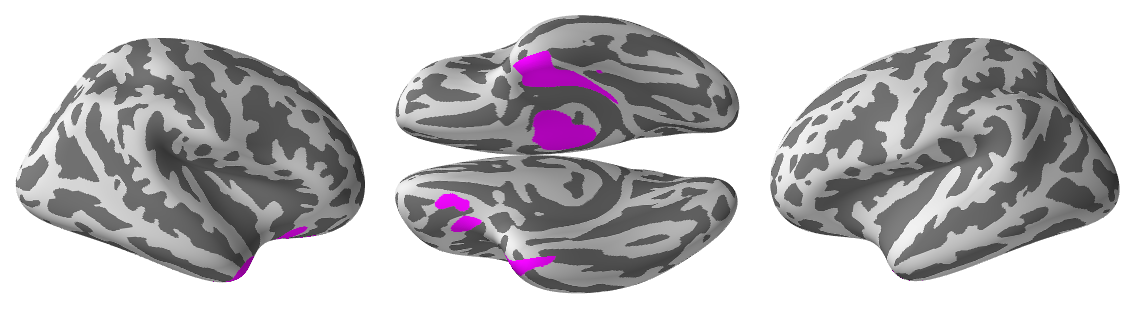}} \\
\includegraphics[width=0.3\textwidth]{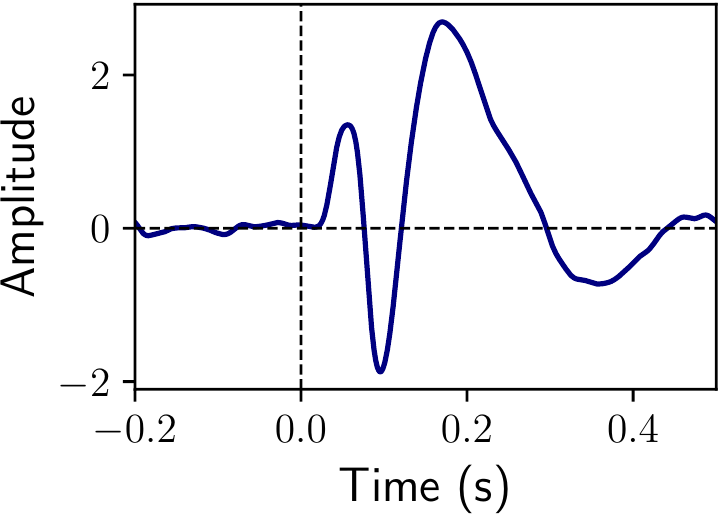}%
\raisebox{0.2\height}{\includegraphics[width=0.68\textwidth]{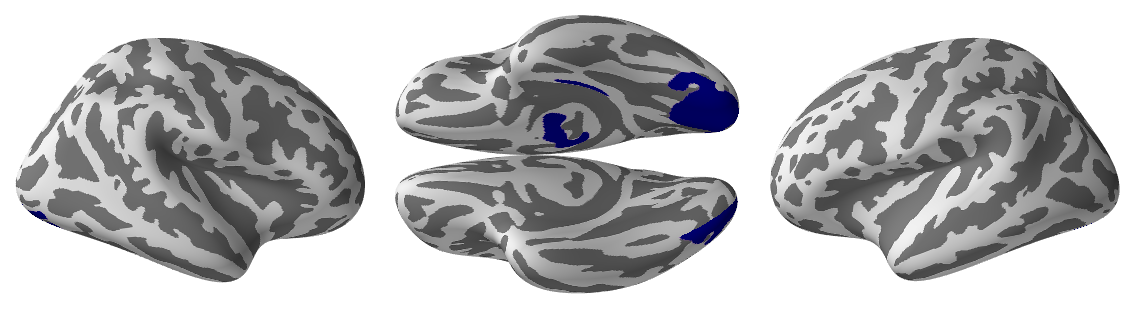}}
\includegraphics[width=0.3\textwidth]{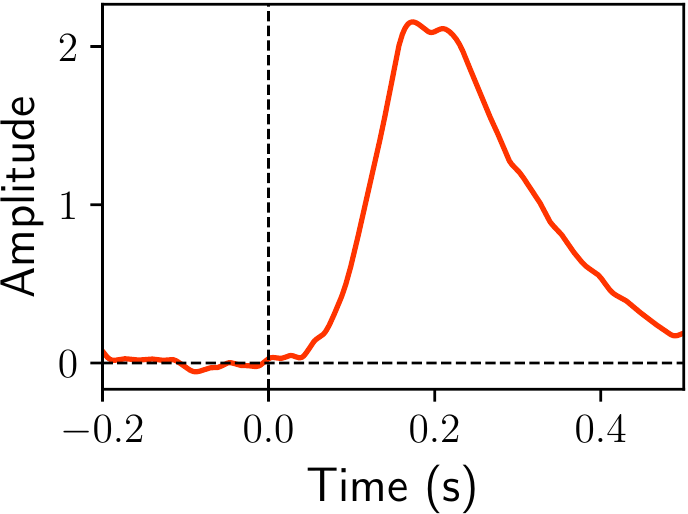}%
\raisebox{0.2\height}{\includegraphics[width=0.68\textwidth]{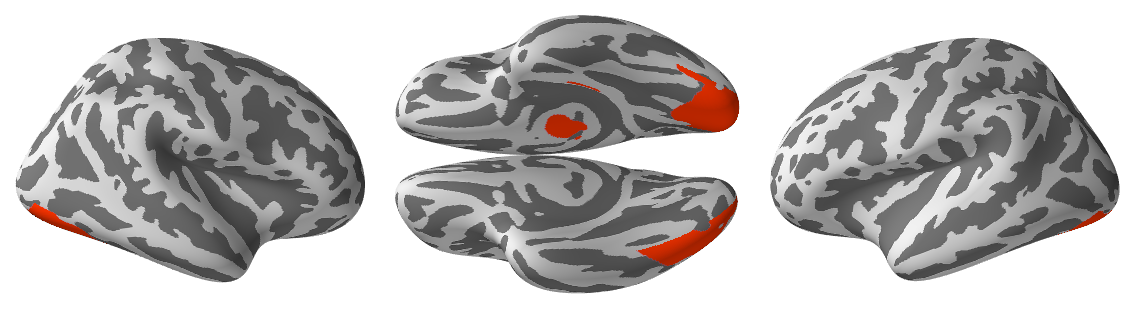}} \\
\includegraphics[width=0.3\textwidth]{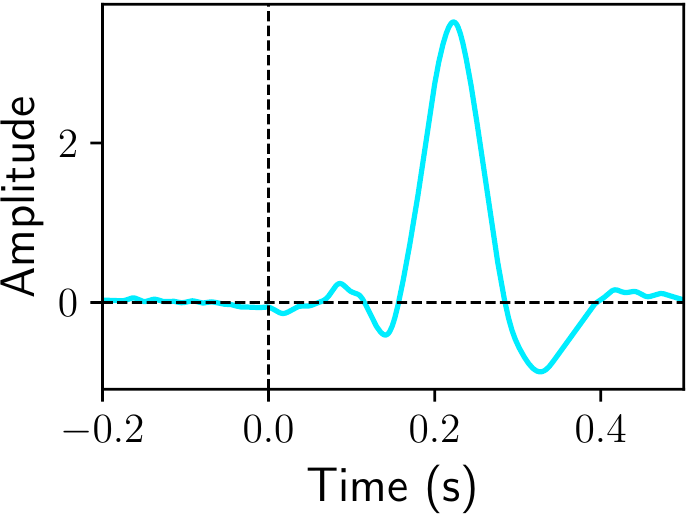}%
\raisebox{0.2\height}{\includegraphics[width=0.68\textwidth]{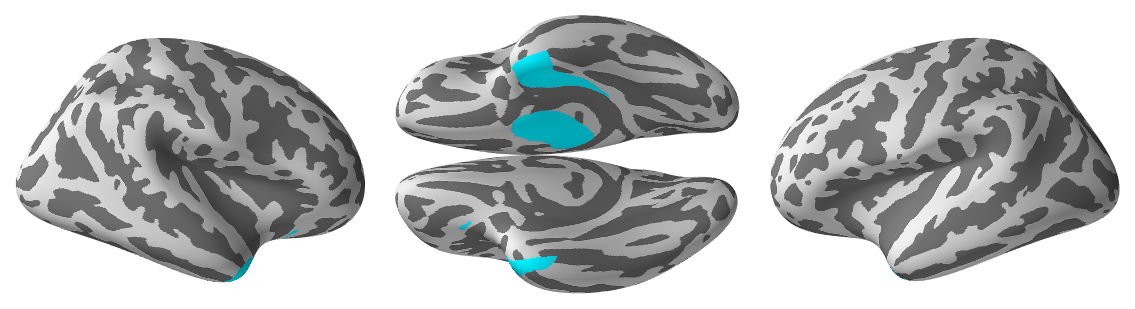}} \\
\includegraphics[width=0.3\textwidth]{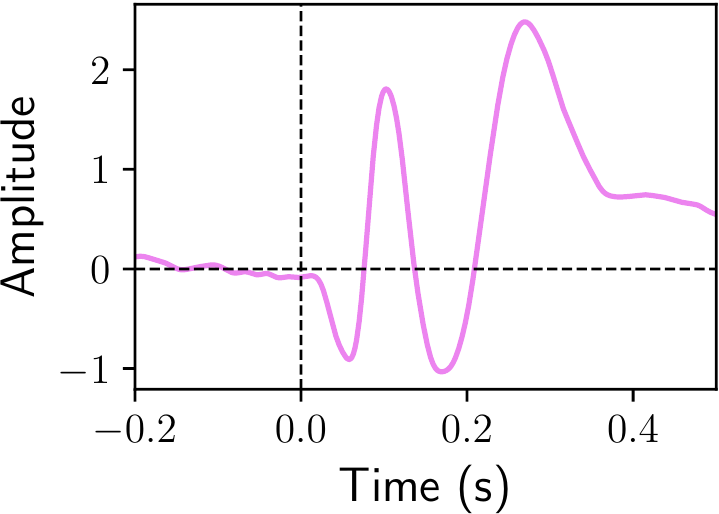}%
\raisebox{0.2\height}{\includegraphics[width=0.68\textwidth]{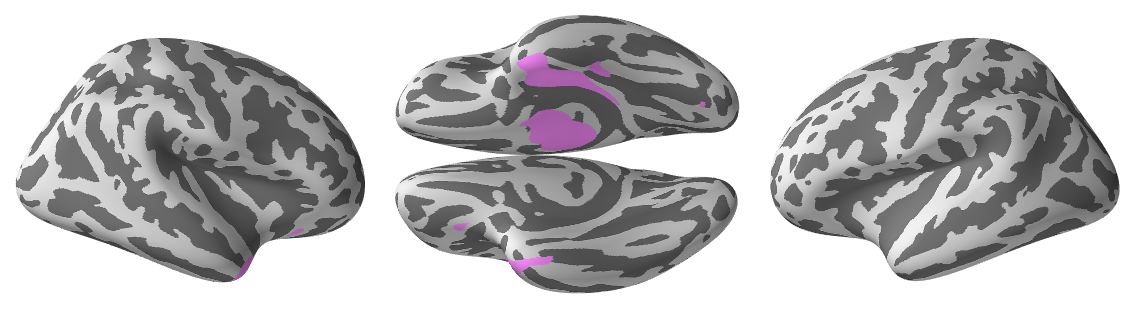}} \\
\includegraphics[width=0.3\textwidth]{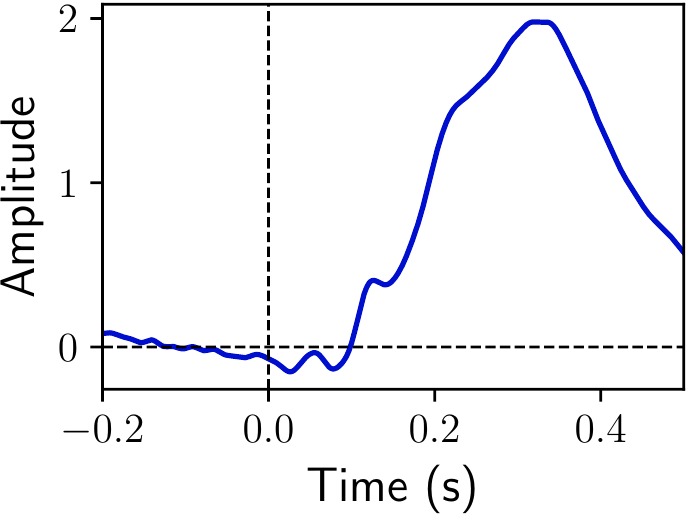}%
\raisebox{0.2\height}{\includegraphics[width=0.68\textwidth]{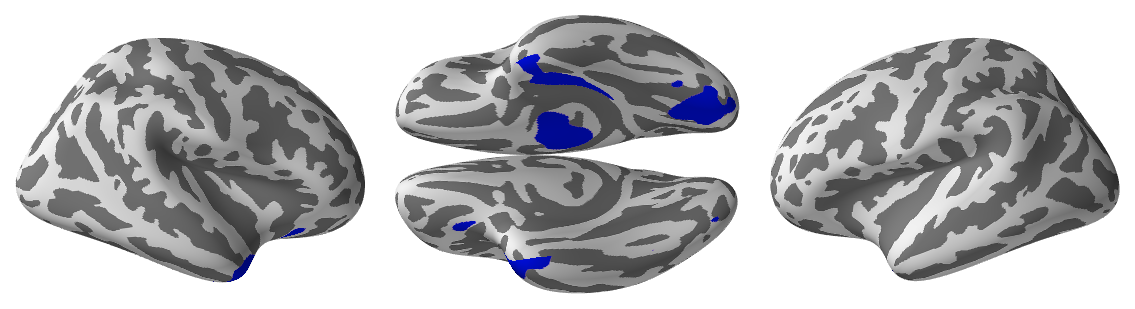}} \\
\includegraphics[width=0.3\textwidth]{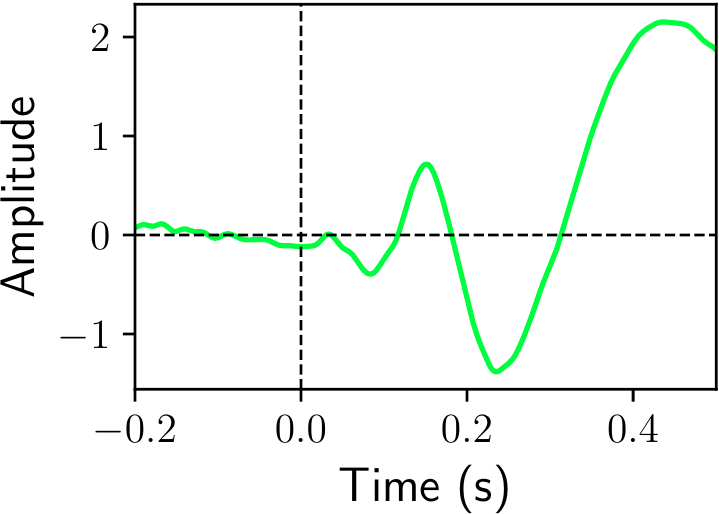}%
\raisebox{0.2\height}{\includegraphics[width=0.68\textwidth]{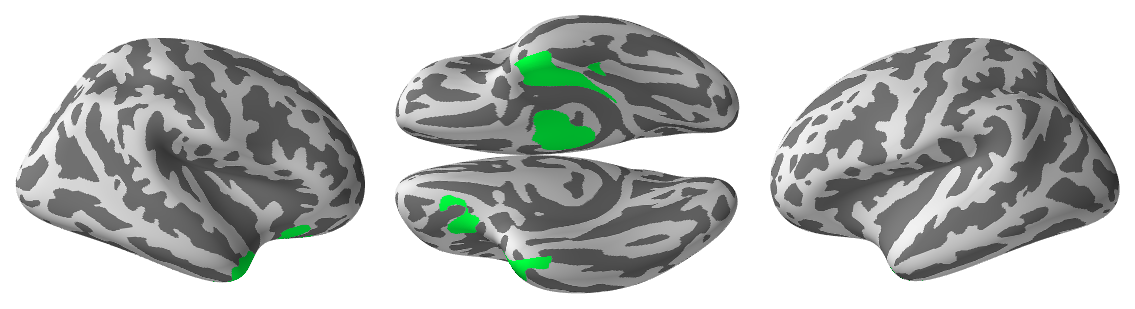}} \\
}

\end{document}